\documentclass[11pt,onecolumn,twoside]{IEEEtran}

\usepackage{amsfonts,amsmath, amssymb, graphicx, cite}

\usepackage{color}
\usepackage{multirow}
\usepackage{rotating}
\usepackage{bbm}
\usepackage{latexsym}

\usepackage{caption,subcaption,comment}
\usepackage{algorithm,algorithmic,amsthm}

\newtheorem{theorem}{Theorem}
\newtheorem{lemma}{Lemma}

\newcommand{\LT}{\Lambda(T)}
\newcommand{\LI}{\Lambda^*}
\newcommand{\LnT}{\Lambda_{\upsilon,\nu}(T)}
\newcommand{\LnI}{\Lambda_{\upsilon,\nu}^*}

\begin{document}

\title{Noise Facilitation in Associative Memories of Exponential Capacity%
\thanks{This work is based in part on a paper presented at the 2013 Neural Information Processing Systems Conference, Lake Tahoe, December 2013 \cite{KarbasiSSV2013}.}
}

\author{Amin Karbasi, Amir Hesam Salavati, Amin Shokrollahi, and Lav~R.~Varshney%
\thanks{A.~Karbasi is with Eidgen{\"{o}}ssische Technische Hochschule Z{\"{u}}rich, Switzerland (e-mail: amin.karbasi@inf.ethz.ch).}
\thanks{A.~H.~Salavati and A.~Shokrollahi are with {\'{E}}cole Polytechnique F{\'{e}}d{\'{e}}rale de Lausanne, Switzerland (e-mail: \{hesam.salavati, amin.shokrollahi\}@epfl.ch).}
\thanks{L.~R.~Varshney was with the IBM Thomas J.\ Watson Research Center, NY.  He is now with the University of Illinois at Urbana-Champaign (e-mail: varshney@illinois.edu).}
}

\maketitle

\begin{abstract}
Recent advances in associative memory design through structured pattern sets and graph-based inference
algorithms have allowed reliable learning and recall of an exponential number of patterns.  Although 
these designs correct external errors in recall, they assume neurons that compute noiselessly, in contrast 
to the highly variable neurons in brain regions thought to operate associatively such as hippocampus 
and olfactory cortex.  

Here we consider associative memories with noisy internal computations and analytically characterize 
performance.  As long as the internal noise level is below a specified threshold, the error probability 
in the recall phase can be made exceedingly small.  More surprisingly, we show that internal noise 
actually improves the performance of the recall phase while the pattern retrieval capacity remains intact, 
i.e., the number of stored patterns does not reduce with noise (up to a threshold).  Computational 
experiments lend additional support to our theoretical analysis.  This work suggests a functional 
benefit to noisy neurons in biological neuronal networks. 
\end{abstract}

\section{Introduction}
\label{sec:introduction}
\IEEEPARstart{B}{rain} regions such as hippocampus and olfactory cortex are thought to operate as 
associative memories \cite{TrevesR1994,StettlerA2009,WilsonS2011}, having the ability to learn patterns 
from presented inputs, store a large number of patterns, and retrieve them reliably in the face of noisy queries 
\cite{Hopfield1982,McEliecePRV1987,AmitF1994}.  Mathematical models of associative memory are therefore
designed to memorize a set of given patterns so that corrupted versions of the memorized patterns 
may later be presented and the correct memorized pattern retrieved.

Although such information storage and recall seemingly falls naturally into the information-theoretic
framework \cite{Palm1980}, where an exponential number of messages can be communicated reliably using a linear number of 
symbols \cite{Shannon1948}, classical associative memory models can only store a linear number of patterns 
with a linear number of symbols \cite{McEliecePRV1987}.  A primary shortcoming of such classical models has 
been their requirement to memorize a randomly chosen set of patterns.  By enforcing structure and redundancy
in the possible set of memorizable patterns---much like natural stimuli \cite{OlshausenF2004}, internal 
representations in neural systems \cite{KoulakovR2011}, and codewords in error-control codes 
\cite{RichardsonU2008}---new advances in associative memory design allow storage of an exponential 
number of patterns with a linear number of symbols \cite{SalavatiK2012,KarbasiSS2013}, just like in 
communication systems.\footnote{The idea of restricted pattern sets leading to associative memories with
increased storage capacity was first suggested in an unpublished doctoral dissertation \cite{Biswas1993}.}

Information-theoretic and associative memory models of storage have been used to predict experimentally 
measurable properties of synapses in the mammalian brain \cite{BrunelHINB2004,VarshneySC2006}.  But 
contrary to the fact that noise is present in computational operations of the brain 
\cite{Koch1999,FaisalSW2008,RollsD2010,McDonnellW2011,DestexheR2012}, associative memory models with exponential 
capacity have assumed no internal noise in the computational nodes \cite{KarbasiSS2013}; likewise with
many classical models \cite{Hopfield1982}.  The purpose of the present paper is to model internal noise 
in associative memories with exponential pattern retrieval capacity and study whether they are still able to operate reliably.  
Surprisingly, we find internal noise actually enhances recall performance without loss in capacity, thereby 
suggesting a functional role for variability in the brain.

In particular we consider a convolutional, graph code-based, associative memory model \cite{KarbasiSS2013} 
and find that even if all components are noisy, the final error probability in recall can be made 
exceedingly small.  We characterize a threshold phenomenon and show how to optimize algorithm parameters 
when knowing statistical properties of internal noise.  Rather counterintuitively the performance of the 
memory model \emph{improves} in the presence of internal neural noise, as has been observed previously
as \emph{stochastic resonance} in the literature \cite{ChenVKM2007,McDonnellW2011}.  Deeper analysis shows
mathematical connections to perturbed simplex algorithms for linear programing \cite{SpielmanT2004}, where 
some internal noise helps the algorithm get out of local minima.

\subsection{Related Work}
\label{sec:related}

Designing neural networks to learn a set of patterns and recall them later in the presence of noise has been 
an active topic of research for the past three decades. Inspired by Hebbian learning \cite{Hebb1949}, Hopfield 
introduced an auto-associative neural mechanism of size $n$ with binary state neurons in which patterns are 
assumed to be binary vectors of length $n$ \cite{Hopfield1982}. The capacity of a Hopfield network under vanishing 
block error probability was later shown to be $O(n/\log(n))$ \cite{McEliecePRV1987}.  With the hope of increasing 
the capacity of the Hopfield network, extensions to non-binary states were explored \cite{AmitF1994}. In particular, 
Jankowski et al.\ investigated a multi-state complex-valued neural associative memory with 
estimated capacity less than $0.15 n$ \cite{JankowskiLZ1996}; Muezzinoglu et al.\ showed the capacity with a 
prohibitively complicated learning rule to increase to $n$ \cite{MuezzinogluGZ2003}. 
Lee proposed the Modified Gradient Descent learning Rule (MGDR) to overcome this drawback \cite{Lee2006}.

To further increase capacity and robustness, a recent line of work considers exploiting structure in patterns. 
This is done either by making use of correlations among patterns or by only memorizing patterns with redundancy
(rather than any possible set of patterns). By utilizing neural cliques, \cite{GriponB2011} demonstrated that 
increasing the pattern retrieval capacity of Hopfield networks to $O(n^2)$ is possible. Modification of neural 
architecture to improve pattern retrieval capacity has also been previously considered by Venkatesh and Biswas 
\cite{Biswas1993, venkatesh_exponential}, where the capacity is increased to $\Theta \left(b^{n/b} \right)$ for 
\emph{semi-random} patterns, where $b = \omega( \ln n)$ is the size of clusters. This significant boost to capacity 
is achieved by dividing the neural network into smaller fully interconnected \emph{disjoint} blocks or \emph{nested} 
blocks (cf.~\cite{Baram1991}). This huge improvement comes at the price of limited \emph{worst-case} noise 
tolerance capabilities. Deploying higher order neural models beyond the pairwise correlation considered in Hopfield 
networks increases the storage capacity to $O(n^{p-2})$, where $p$ is the degree of correlation \cite{PerettoN1986}. 
In such models, neuronal state depends not only on the state of neighbors, but also on the correlations among them. 
A new model based on bipartite graphs that captures higher-order correlations (when patterns belong to a subspace), 
but without prohibitive computational complexity, improved capacity to $O(a^n)$, for some $a > 1$, that is to say 
\emph{exponential} in network size \cite{SalavatiK2012}. 

The basic memory architecture, learning rule, and recall algorithm used herein is from \cite{KarbasiSS2013}, 
which also achieves exponential capacity by capturing internal redundancy by dividing the patterns into smaller 
\emph{overlapping} clusters, with each subpattern satisfying a set of linear constraints. The problem of learning 
linear constraints with neural networks was considered in \cite{XuKO1991}, but without sparsity requirements.  
This has connections to compressed sensing \cite{CandesT2006}; typical compressed sensing recall/decoding algorithms 
are too complicated to be implemented by neural networks, but some have suggested the biological plausibility of 
message-passing algorithms \cite{FletcherRVB2011}.

Building on the idea of structured pattern sets \cite{GriponB2011}, the basic associative memory model used 
herein \cite{KarbasiSS2013} relies on the fact all patterns to be learned lie in a low-dimensional subspace.  
Learning features of a low-dimensional space is very similar to autoencoders \cite{VincentLBM2008}. 
The model also has similarities to Deep Belief Networks (DBNs) and in particular Convolutional Neural Networks
\cite{LeNCCKN2010}, albeit with different objectives.  DBNs are made of several consecutive stages, similar 
to overlapping clusters in our model, where each stage extracts some features and feeds them to the following stage. 
The output of the last stage is then used for pattern classification. In contrast to DBNs, our associative memory 
model is not classifying patterns but rather recalling patterns from noisy versions.  Also, overlapping clusters 
can operate in parallel to save time in information diffusion over a staged architecture.

In most deep or convolutional models, one not only has to find the proper dictionary for classification, 
but also calculate the features for each input pattern. This increases the complexity of the whole system when
the objective is simply recall.  Here the dictionary corresponds to the dual vectors from previously memorized patterns.

In this work, we reconsider the neural network model of \cite{KarbasiSS2013}, but introduce internal computation 
noise consistent with biology.  Note that the sparsity of the model architecture is also consistent with 
biology \cite{SongSRNC2005}.  We find that there is actually a functional benefit to internal noise.

The benefit of internal noise has been noted previously in associative memory models with stochastic update 
rules, cf.~\cite{Amit1992}, by analyzing attractor dynamics. In particular, it has been shown that noise may 
reduce recall time in associative memory tasks by pushing the system from one attractor state to another 
\cite{LiljenstromW1995}.  However, our framework differs from previous approaches in three key aspects. First, 
our memory model is different, which makes extension of previous analysis nontrivial.  Second, and perhaps most 
importantly, pattern retrieval capacity in previous approaches \emph{decreases} with internal noise, 
cf.~\cite[Figure~6.1]{Amit1992}, in that increasing internal noise helps correct more external errors, but also 
reduces the number of memorizable patterns. In our framework, internal noise does not affect pattern retrieval 
capacity (up to a threshold) but improves recall performance.  Finally, our noise model has bounded rather 
than Gaussian noise, and so a suitable network may achieve \emph{perfect} recall despite internal noise.

Reliably storing information in memory systems constructed completely from unreliable components is a 
classical problem in fault-tolerant computing \cite{Taylor1968,Kuznetsov1973,Varshney2011}, where typical 
models have used random access architectures with sequential correcting networks.  Although direct comparison 
is difficult since notions of circuit complexity are slightly different, our work also demonstrates that 
associative memory architectures can store information reliably despite being constructed from unreliable 
components.

\section{Associative Memory Model}
\label{sec:model}  
In this section, we introduce  our main notation, the model of associative memories and noise. We also explain the recall algorithms. 
\subsection{Notation and basic structure}
In our model, a neuron  can assume an integer-valued state from the set $\mathcal{Q} = \{0,\dots,Q-1\}$, interpreted 
as the short term firing rate of neurons. A neuron updates its state based on the states of its neighbor 
$\{s_i\}_{i=1}^{n}$ as follows. It first computes a weighted sum $h = \sum_{i=1}^{n} w_i s_i + \zeta$, where 
$w_i$ is the weight of the link from $s_i$ and $\zeta$ is the \emph{internal noise}, and then applies nonlinear 
function $f: \mathbb{R} \rightarrow \mathcal{Q}$ to $h$. 

\begin{figure}
  \centering
  \begin{subfigure}[b]{0.5\textwidth}
    \includegraphics[width=0.99\textwidth]{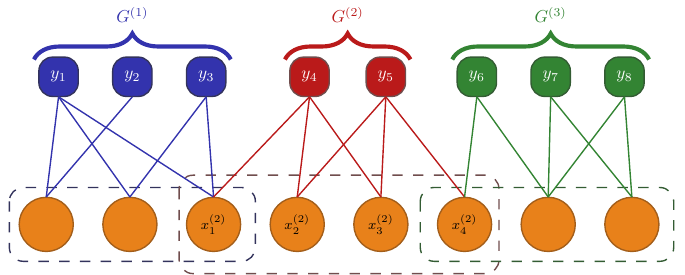}
    \caption{Bipartite graph $G$.}
    \label{overlapping_clustered_network}
  \end{subfigure}
  \quad
  \begin{subfigure}[b]{0.5\textwidth}
    \includegraphics[width=0.99\textwidth]{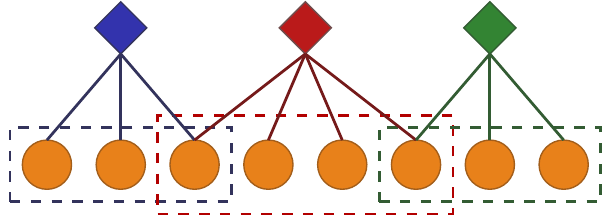}
    \caption{Contraction graph $\widetilde{G}$.}
    \label{fig:inter}
  \end{subfigure}
\caption{The proposed neural associative memory with overlapping clusters.}
\label{fig_model}
\end{figure}

An associative memory is represented by a weighted bipartite graph, $G$, with pattern neurons and constraint 
neurons.  Each pattern $x = (x_1, x_2, \ldots, x_n)$ is a vector of length $n$, where $x_i\in \mathcal{Q}$, 
$i = 1,\ldots,n$. Following \cite{KarbasiSS2013}, the focus is on recalling patterns with strong 
\emph{local correlation} among entries. Hence, we divide entries of each pattern $x$ into $L$ \emph{overlapping} 
subpatterns of lengths $n_1, n_2, \ldots, n_L$. Due to overlaps, a pattern neuron can be a member of multiple 
subpatterns, as depicted in Figure~\ref{overlapping_clustered_network}. The $i$th subpattern is denoted 
$x^{(i)} = (x_1^{(i)}, x_2^{(i)}, \ldots, x_{n_i}^{(i)})$, and local correlations are assumed to be in the form 
of subspaces, i.e.\ the subpatterns $x^{(i)}$ form a subspace of dimension $k_i<n_i$.

We capture the local correlations by learning a set of linear constraints over each subspace corresponding to the 
dual vectors orthogonal to that subspace. More specifically, let $\{w^{(i)}_1, w^{(i)}_2, \ldots, w^{(i)}_{m_i}\}$ 
be a set of dual vectors orthogonal to all subpatterns $x^{(i)}$ of cluster $i$. Then:
\begin{equation}
\label{eq:orthogonal}
y_j^{(i)}=(w_j^{(i)})^T\cdot x^{(i)}=0,  \quad \mbox{for all } j\in \{1,\ldots,m_i\} \mbox{ and for all } i\in \{1,\ldots,L\}\mbox{.}
\end{equation}
Eq.~\eqref{eq:orthogonal} can be rewritten as $W^{(i)}\cdot x^{(i)} = 0$ where $W^{(i)} = [w_1^{(i)}| w_2^{(i)}| \cdots| w_{m_i}^{(i)}]^T$
 is the matrix of dual vectors. Now we use a bipartite graph with connectivity matrix determined by $W^{(i)}$ to 
represent the subspace constraints learned from subpattern $x^{(i)}$; this graph is called \emph{cluster} $i$. We 
developed an efficient way of learning $W^{(i)}$ in \cite{KarbasiSS2013}, also used here. Briefly, in each iteration 
of learning:
\begin{enumerate}
\item Pick a pattern $x$ at random from the dataset;
\item Adjust weight vectors $w_j^{(i)}$ for $j = \{1,\dots,m_i\}$ and $i = \{1,\dots,L\}$ such that the projection 
of $x$ onto $w_j^{(i)}$ is reduced. Apply a sparsity penalty to favor sparse solutions.
\end{enumerate}
This process repeats until all weights are orthogonal to the patterns in the dataset or the maximum iteration limit 
is reached.  The learning rule allows us to assume the weight matrices $W^{(i)}$ are known and satisfy 
$W^{(i)}\cdot x^{(i)} = 0$ for all patterns $x$ in the dataset $\mathcal{X}$, in this paper.

For the forthcoming asymptotic analysis, we need to define a \emph{contracted graph} $\widetilde{G}$ whose 
connectivity matrix is denoted $\widetilde{W}$ and has size $L \times n$. This is a bipartite graph in which 
constraints in each cluster are represented by a single neuron. Thus, if pattern neuron $x_j$ is connected to 
cluster $i$, $\widetilde{W}_{ij} = 1$; otherwise $\widetilde{W}_{ij} = 0$, see Figure~\ref{fig:inter}. 
We also define the degree distribution from an \emph{edge perspective} over $\widetilde{G}$, using
\begin{align}
\label{edge-degree}
\widetilde{\lambda}(z) &= \sum_j \widetilde{\lambda}_j z^{j-1} \mbox{, and } \\
\widetilde{\rho}(z) &= \sum_j \widetilde{\rho}_j z^{j-1}\mbox{,}
\end{align}
where $\widetilde{\lambda}_j$ (resp., $\widetilde{\rho}_j$) equals the fraction of edges that connect to 
pattern (resp., cluster) nodes of degree $j$. 

\subsection{Noise model}
There are two types of noise in our model: \emph{external errors} and \emph{internal noise}. As mentioned 
earlier, a neural network should be able to retrieve memorized pattern $\hat{x}$ from its corrupted version 
$x$ due to external errors. We assume the external error is an additive vector of size $n$, denoted by $z$ 
satisfying $x=\hat{x}+z$, whose entries assume values independently from $\{-1,0,+1\}$\footnote{Note that 
the proposed algorithms also work with larger noise values, i.e. from a set $\{-S,\dots,S\}$ for some 
$S\in \mathbb{N}$, see Sec.~\ref{sec:larger}; the $\pm 1$ noise model is presented here for simplicity.} 
with corresponding probabilities $p_{-1} =p_{+1}=\epsilon/2$ and $p_0=1-\epsilon$. The realization of the 
external error on subpattern $x^{(i)}$ is denoted $z^{(i)}$.  Note that the subspace assumption implies 
$W\cdot y = W\cdot z$ and $W^{(i)} \cdot y^{(i)} = W^{(i)} \cdot z^{(i)}$ for all $i$.

Neurons also suffer from internal noise. We consider a bounded noise model, i.e.\ a random number uniformly 
distributed in the intervals $[-\upsilon,\upsilon]$ and $[-\nu,\nu]$ for the pattern and constraint neurons, 
respectively ($\upsilon,\nu < 1$).

The goal of recall is to filter the external error $z$ to obtain the desired pattern $x$ as the correct 
states of the pattern neurons. When neurons compute noiselessly, this task may be achieved by exploiting the 
fact the set of patterns $x \in\mathcal{X}$ satisfy the set of constraints $W^{(i)} \cdot x^{(i)} = 0$. 
However, it is not clear how to accomplish this objective when the neural computations are noisy. Rather 
surprisingly, we show that eliminating external errors is not only possible in the presence of  internal 
noise, but that neural networks with moderate internal noise demonstrate better external error resilience.

\subsection{Recall algorithms}
To efficiently deal with external errors in associative memory, we use two simple iterative 
message passing algorithms. The role of the first one, called the \emph{Intra-cluster} algorithm 
and formally defined in Algorithm~\ref{algo:correction}, is to correct at least a single external 
error in each cluster. However, without overlaps between clusters, the error resilience of this 
approach and the network in general is limited. The second algorithm, the \emph{Inter-cluster} 
recall algorithm, exploits the overlaps: it helps clusters with external errors recover their 
correct states by using the reliable information from clusters that do not have external errors. 
The error resilience of the resulting combination thereby drastically improves.

To go further into details, and with abuse of notations, let $x_i(t)$ and $y_j(t)$ denote the message 
transmitted at iteration $t$ by pattern and constraint neurons, respectively. In the first iteration, 
we initialize the pattern neurons with a pattern randomly drawn from the dataset, $\hat{x}$, corrupted 
with some external noise, $z$. Thus, $x(0) = \hat{x} + z$. As a result, for cluster $\ell$ we have 
$x^{(\ell)}(0) = \hat{x}^{(\ell)} + z^{(\ell)}$, where $z^{(\ell)} $ is the realization of the external 
error on cluster $\ell$.

With these notations in mind, Algorithm~\ref{algo:correction} iteratively performs a series of forward 
and backward steps in order to remove (at least) one external error from its input domain. Assuming that 
the algorithm is applied to cluster $\ell$, in the forward step of iteration $t$ the pattern neurons 
in cluster $\ell$ transmit their current states to their neighboring constraint neurons. Each constraint 
neuron $j$ then calculates the weighted sum of the messages it received over its input links. Nevertheless, 
since neurons suffer from internal noise, additional noise terms appear in the weighted sum, i.e., 
$h^{(\ell)}_j = \sum_{i=1}^{n_\ell} W^{(\ell)}_{ij} x^{(\ell)}_i + v_i$, where $v_i$ is the random 
internal noise affecting node $i$. As before, we consider a bounded noise 
model for $v_i$, i.e., it is uniformly distributed in the interval $[-\nu,\nu]$ for some $\nu < 1$.

A non-zero input sum, excluding the effect of $v_i$, is an indication of the presence of external 
errors among the pattern neurons. Thus, constraint neurons set to their states to the sign of the 
received weighted sum if its magnitude is larger than a fixed threshold, $\psi$. More specifically, 
constraint neuron $j$ updates its state based on the received weighted sum according to the following rule
\begin{equation}
\label{update_rule_constraint}
y^{(\ell)}_j(t) = f(h^{(\ell)}_j(t),\psi) = \begin{cases} +1, & \mbox{if } h^{(\ell)}_j(t) \geq \psi
\\
0, & \mbox{if } -\psi \leq h^{(\ell)}_j(t) \leq \psi
\\
-1, & \mbox{otherwise,}
\end{cases}
\end{equation}
Here, $x^{(\ell)}(t)=[x^{(\ell)}_1(t),\dots,x^{(\ell)}_{n_\ell}(t)]$ is the vector of messages transmitted by 
the pattern neurons and $v_i$ is the random internal noise affecting node $i$.\footnote{Note 
that although the values of $y^{(\ell)}_i(t)$ can be shifted to $0,1,2$, instead of $-1,0,1$ to match our assumption 
that neural states are non-negative, we leave them as such to simplify later analysis.}

In the backward step, the constraint neurons communicate their states to their neighboring pattern neurons. 
The pattern neurons then compute a normalized weighted sum on the messages they receive over their input link 
and update their current state if the amount of received (non-zero) feedback exceeds a threshold. Otherwise, 
they will retain their current state for the next round. More specifically, pattern node $i$ in cluster $\ell$ 
updates its state in round $t$ according to the equation below
\begin{equation}
\label{update_rule_pattern}
x^{(\ell)}_i(t+1) = \begin{cases} x^{(\ell)}_i(t)-\mbox{sign}(g^{(\ell)}_i(t)), & \mbox{if } |g^{(\ell)}_i(t)| \geq \varphi 
\\ x^{(\ell)}_i(t), & \mbox{otherwise,}
\end{cases}
\end{equation}
where $\varphi$ is the update threshold and 
\[
g^{(\ell)}_i(t) = \frac{\left((\mbox{sign}(W^{(\ell)})^\top\cdot y^{(\ell)}(t)\right)_i}{d^{(\ell)}_i} + u_i\mbox{.}
\]
Note that $x^{(\ell)}_i(t+1)$ is further mapped to the interval $[0,Q-1]$ by saturating the values below $0$ 
and above $Q-1$ to $0$ and $Q-1$ respectively; this saturation is not stated mathematically for 
brevity. Here, $d^{(\ell)}_i$ is the degree of pattern node $i$ in cluster $\ell$, 
$y^{(\ell)}(t)=[y^{(\ell)}_1(t),\ldots,y^{(\ell)}_{m_\ell}(t)]$ is the vector of messages transmitted by the 
constraint neurons in cluster $\ell$, and $u_i$ is the random internal noise affecting pattern node $i$. Basically, 
the term $g^{(\ell)}_i(t)$ reflects the (average) belief of constraint nodes connected to pattern neuron 
$i$ about its correct value. If $g^{(\ell)}_i(t)$ is larger than a specified threshold $\varphi$ it means most 
of the connected constraints suggest the current state $x^{(\ell)}_i(t)$ is not correct, hence, a change 
should be made. Note this average belief is diluted by the internal noise of neuron $i$.   As mentioned earlier, 
$u_i$ is uniformly distributed in the interval $[-\upsilon,\upsilon]$, for some $\upsilon < 1$. 

\begin{figure}[t]
\begin{algorithm}[H]
\caption{Intra-Module Error Correction} 
\label{algo:correction}
\begin{algorithmic}[1]
\REQUIRE{Training set $\mathcal{X}$, thresholds $\varphi, \psi$, iteration $t_{\max}$}
\ENSURE{$x^{(\ell)}_1,x^{(\ell)}_2,\dots,x^{(\ell)}_{n_\ell}$}
\FOR{$t = 1 \to t_{\max}$} 
\STATE \emph{Forward iteration:} Calculate the input $ h_i^{(\ell)} = \sum_{j=1}^{n_\ell} W^{(\ell)}_{ij} x^{(\ell)}_j + v_i,$ for each neuron $y^{(\ell)}_i$ and set
$y^{(\ell)}_i = f(h^{(\ell)}_i,\psi).$
\STATE \textit{Backward iteration:} Each neuron $x^{(\ell)}_j$ computes\begin{center} $g^{(\ell)}_j = \frac{\sum_{i=1}^{m_\ell} \mbox{sign}(W^{(\ell)}_{ij}) y^{(\ell)}_i}{\sum_{i = 1}^{m_\ell}\mbox{sign}(|W^{(\ell)}_{ij}|)}+u_i.$ \end{center}
\STATE Update the state of each pattern neuron $j$ according to $x^{(\ell)}_j = x^{(\ell)}_j - \hbox{sign}(g^{(\ell)}_j)$
only if $|g^{(\ell)}_j| > \varphi$.
\ENDFOR
\end{algorithmic}
\end{algorithm}
\end{figure}

\begin{figure}
\begin{algorithm}[H]
\caption{Sequential Peeling Algorithm}
\label{algo:peeling}
\begin{algorithmic}[1]
\REQUIRE{$\widetilde{G}, G^{(1)}, G^{(2)}, \dots, G^{(L)}$.}
\ENSURE{$x_1,x_2,\dots,x_n$}
\WHILE{there is an unsatisfied $v^{(\ell)}$}
\FOR{$\ell = 1 \to L$} 
\STATE If $v^{(\ell)}$ is unsatisfied, apply Algorithm~\ref{algo:correction} to cluster $G^{(l)}$.
\STATE If $v^{(\ell)}$ remained unsatisfied, revert the state of pattern neurons connected to $v^{(\ell)}$ to their initial state. Otherwise, keep their current states.
\ENDFOR
\ENDWHILE
\STATE Declare {$x_1,x_2,\dots,x_n$} if all $v^{(\ell)}$'s are satisfied. Otherwise, declare failure. 
\end{algorithmic}
\end{algorithm} 
\end{figure}

The error correction ability of Algorithm~\ref{algo:correction} is fairly limited, as determined analytically and through 
simulations in the sequel.  In essence, Algorithm~\ref{algo:correction} can correct one external error with high probability, 
but degrades terribly against two or more external errors. Working independently, clusters cannot correct more than a 
few external errors, but their combined performance is much better. As clusters overlap, they help each other in resolving 
external errors: a cluster whose pattern neurons are in their correct states can \emph{always} provide truthful information 
to neighboring clusters. This property is exploited in Algorithm~\ref{algo:peeling} by applying Algorithm~\ref{algo:correction} 
in a round-robin fashion to each cluster. Clusters either eliminate their internal noise in which case they keep their 
new states and can now help other clusters, or revert back to their original states. Note that by such a scheduling scheme, 
neurons can only change their states towards correct values. This scheduling technique is similar in spirit to the peeling 
algorithm \cite{LubyMSS2001a}. 

\section{Pattern Retrieval Capacity}
\label{sec:nosiy_capacity}
Before proceeding to analyze recall performance, for completeness we review pattern retrieval capacity results 
from \cite{KarbasiSS2013} to show that the proposed model is capable of memorizing an exponentially large number 
of patterns. First, note that since the patterns form a subspace, the number of patterns $C$ does not have any effect 
on the learning or recall algorithms (except for its obvious influence on the learning time). Thus, in order to show 
that the pattern retrieval capacity is exponential in $n$, all we need to demonstrate is that there exists a training 
set $\mathcal{X}$ with $C$ patterns of length $n$ for which $C \propto a^{rn}$, for some $a > 1$ and $0<r$. 
\begin{theorem}[\cite{KarbasiSS2013}]
\label{theorem_exponential_solution}
Let $\mathcal{X}$ be a $\mathcal{C} \times n$ matrix, formed by $\mathcal{C}$ vectors of length $n$ with entries 
from the set $\mathcal{Q}$. Furthermore, let $k = rn$ for some $0<r<1$. Then, there exists a set of vectors for 
which $\mathcal{C} = a^{rn}$, with $a > 1$, and $\mbox{rank}(\mathcal{X}) = k <n$.
\end{theorem}
The proof is constructive: we create a dataset $\mathcal{X}$ such that it can be memorized by the proposed neural 
network and satisfies the required properties, i.e.\ the subpatterns form a subspace and pattern entries are integer 
values from the set $\mathcal{Q} = \{0,\dots,Q-1\}$. The complete proof can be found in \cite{KarbasiSS2013}.

\section{Recall Performance Analysis}
\label{sec:main}

Now let us analyze recall error performance.  The following lemma shows that if $\varphi$ and $\psi$ are chosen 
properly, then in the absence of external errors the constraints remain satisfied and internal noise cannot result 
in violations. This is a crucial property for Algorithm~\ref{algo:peeling}, as it allows one to determine whether a 
cluster has successfully eliminated external errors (Step 4 of algorithm) by merely checking the satisfaction of 
all constraint nodes.
\begin{lemma}
\label{lem:pi_0}
In the absence of external errors, the probability that a constraint neuron (resp. pattern neuron) in cluster 
$\ell$ makes a wrong decision due to its internal noise is given by $\pi^{(\ell)}_0 = \max \left(0,\frac{\nu-\psi}{\nu}\right)$ 
(resp. $P^{(\ell)}_0 = \max \left(0,\frac{\upsilon-\varphi}{\upsilon}\right)$). 
\end{lemma}
\begin{IEEEproof}
To calculate the probability that a constraint node makes a mistake when there are no external errors, consider 
constraint node $i$ whose decision parameter will be
\[
h^{(\ell)}_i = \left(W^{(\ell)}\cdot x^{(\ell)}\right)_i + v_i = v_i \mbox{.}
\]
Therefore, the probability of making a mistake will be
\begin{equation}
\label{Pi_0_original}
\pi^{(\ell)}_0 = \mbox{Pr}\{ |v_i| > \psi \} = \max \left(0,\frac{\nu-\psi}{\nu}\right)\mbox{.}
\end{equation}
Thus, to make $\pi^{(\ell)}_0=0$ we will select $\psi > \nu$. Note that this might not be possible in all cases since, 
as we will see, the minimum absolute value of network weights should be at least $\psi$; if $\psi$ is too large we 
might not be able to find a proper set of weights. Nevertheless, and assuming that it is possible to choose a proper 
$\psi$, we will have
\begin{equation}
\label{Pi_0}
\pi^{(0)} = 0 \mbox{.}
\end{equation}

Now knowing that the constraint will not send any non-zero messages in the absence of external noise, 
we focus on the pattern neurons in the same circumstance. A given pattern node $x^{(\ell)}_j$ will receive 
a zero from all its neighbors among the constraint nodes. Therefore, its decision parameter will be 
$g^{(\ell)}_j = u_j$. As a result, a mistake could happen if $|u_j| \geq \varphi$. The probability of this 
event is given by
\begin{equation}
\label{P_1_0}
P^{(\ell)}_0 = \mbox{Pr}\{ |u_i| > \varphi \} = \max \left(0,\frac{\upsilon-\varphi}{\varphi}\right)\mbox{.}
\end{equation}
Therefore, to make $P^{(\ell)}_0$ go to zero, we must select $\varphi \geq \upsilon$. 
\end{IEEEproof}

In the sequel, we assume $\varphi > \upsilon$ and $\psi > \nu$ so that $\pi^{(\ell)}_0=0$ and 
$P^{(\ell)}_0=0$. However, an external error combined with internal noise may still push neurons 
to an incorrect state.

Given the above lemma and our neural architecture, we can prove the following surprising result: 
in the asymptotic regime of increasing number of iterations of Algorithm~\ref{algo:peeling}, a neural 
network with internal noise outperforms one without, with the pattern retrieval capacity remaining intact. 
Let us define the fraction of errors corrected 
by the noiseless and noisy neural network (parametrized by $\upsilon$ and $\nu$) after $T$ iterations 
of Algorithm~\ref{algo:peeling} by $\LT$ and $\LnT$, respectively. Note that both $\LT\leq 1$ and $\LnT\leq 1$ 
are non-decreasing sequences of $T$. Hence, their limiting values are well defined: 
$\lim_{T\rightarrow\infty} \LT = \LI$ and $\lim_{T\rightarrow\infty} \LnT = \LnI$.

\begin{figure}
 \centering
 \includegraphics[width=5in]{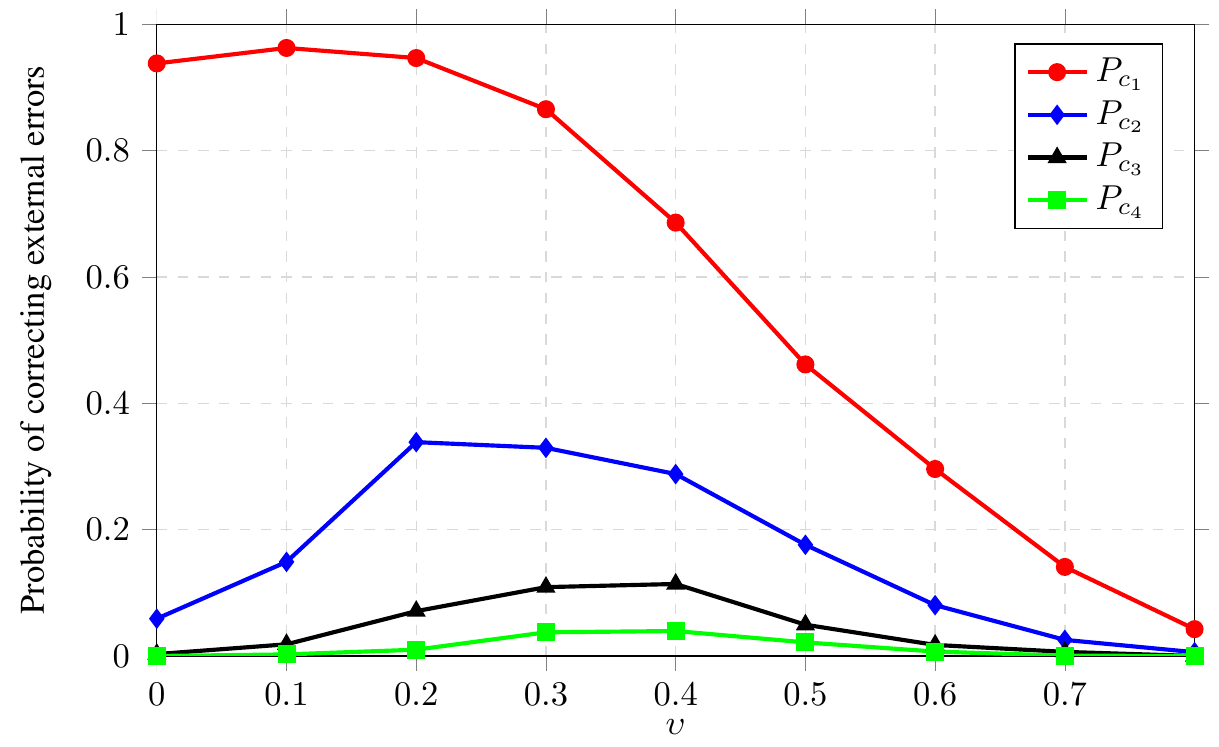}
 \caption{The value of $P_{c_i}$ as a function of pattern neurons noise $\upsilon$ for $i=1,\dots,4$. The noise at constraint neurons is assumed to be zero ($\nu = 0$).} 
 \label{fig:P_c_i_calculation}
\end{figure}

\begin{theorem}
\label{th:noisy_decoder_as_good_as}
Let us choose $\varphi$ and $\psi$ so that $\pi^{(\ell)}_0=0$ and $P^{(\ell)}_0=0$ for all 
$\ell\in\{1,\ldots,L\}$. For the same realization of external errors, we have $\LnI\geq \LI$. 
\end{theorem}
\begin{IEEEproof}
We first show that the noisy network can correct any external error pattern that the noiseless counterpart can 
correct in the $T \rightarrow \infty$ limit. If the noiseless decoder succeeds, then there is a non-zero probability 
$P$ that the noisy decoder succeeds in a given round as well (corresponding to the case that noise values are 
rather small). Since we do not introduce new errors during the application of Algorithm~\ref{algo:peeling}, the number 
of errors in the new rounds are smaller than or equal to the previous round, hence the probability of success 
is lower bounded by $P$. If Algorithm~\ref{algo:peeling} is applied $T$ times, then the probability of correcting the 
external errors at the end of round $T$ is $P + P(1-P) + \cdots + P(1-P)^{T-1} = 1-(1-P)^T$. Since $P > 0$, 
for $T \rightarrow \infty$ this probability tends to $1$. 

Now, we turn attention to cases where the noiseless network fails in eliminating external errors and show that 
there exist external error patterns, called \emph{stopping sets}, for which the noisy network is capable of 
eliminating them while the noiseless network has failed; see Appendix~\ref{sec:proof_thm_noisy_decoder_as_good_as}
for further explication. Assuming that each cluster can eliminate $i$ external 
errors in their domain and in the absence of internal noise,\footnote{From the forthcoming 
Figure~\ref{fig:P_c_i_calculation}, we will see that $i=2$ in this case.} stopping sets correspond to noise patterns 
in which each cluster has more than $i$ errors.  Then Algorithm~\ref{algo:peeling} cannot proceed any further. However, 
in the noisy network, there is a chance that in one of the rounds, the noise acts in favorably and the cluster could 
correct more than $i$ errors.\footnote{This is reflected in the forthcoming Figure~\ref{fig:P_c_i_calculation}, where 
the value of $P_{c_i}$ is larger when the network is noisy.} In this case, if the probability of getting out of the 
stopping set is $P$ in each round, for some $P > 0$, then a similar argument to the previous case shows that 
$P \rightarrow 1$ when $T \rightarrow \infty$.
\end{IEEEproof}
It should be noted that if the amount of internal noise or external errors is too high, the noisy architecture will 
eventually get stuck just like the noiseless network. The high level idea why a noisy network outperforms a noiseless 
one comes from understanding stopping sets, realizations of external errors where the iterative Algorithm~\ref{algo:peeling} 
cannot correct them all.  We showed that the stopping set shrinks as we add internal noise and so the supposedly harmful 
internal noise helps Algorithm~\ref{algo:peeling} to avoid stopping sets.  Appendix~\ref{sec:proof_thm_noisy_decoder_as_good_as}
illustrates this notion further.

Theorem~\ref{th:noisy_decoder_as_good_as} suggests the only possible downside to using a noisy network is 
its possible running time in eliminating external errors: the noisy neural network may need more iterations 
to achieve the same error correction performance. Interestingly, our empirical experiments show that in 
certain scenarios, even the running time improves when using a noisy network. 

Theorem~\ref{algo:peeling} indicates that noisy neural networks (under our model) outperform noiseless ones, but does not 
specify the level of errors that such networks can correct. Now we derive a theoretical upper bound on error correction 
performance. To this end, let $P_{c_i}$ be the average probability that a cluster can correct $i$ external errors in its 
domain. The following theorem gives a simple condition under which Algorithm~\ref{algo:peeling} can correct a linear fraction 
of external errors (in terms of $n$) with high probability. The condition involves $\tilde{\lambda}$ and $\tilde{\rho}$, 
the degree distributions of the contracted graph $\tilde{G}$. 
\begin{theorem}
\label{th:outside}
Under the assumptions that graph $\widetilde{G}$ grows large and it is chosen randomly with degree distributions given 
by $\widetilde{\lambda}$ and $\widetilde{\rho}$, Algorithm~\ref{algo:peeling} is successful if 
\begin{equation}
\label{eq:theorem_main_result}
\epsilon \widetilde{\lambda}\left(1-\sum_{i \geq 1} P_{c_i} \frac{z^{i-1}}{i!} \cdot \frac{d^{i-1}\widetilde{\rho}(1-z)}{dz^{i-1}}\right)<z,\ for\ z\in [0,\epsilon]. 
\end{equation}
\end{theorem}
\begin{IEEEproof}
The proof is based on the density evolution technique \cite{RichardsonU2008}. Without loss of generality, assume we 
have $P_{c_1}$, $P_{c_2}$, and $P_{c_3}$ (and $P_{c_i} = 0$ for $i > 3$.) but the proof can easily be extended if we 
have $P_{c_i}$ for $i>3$. Let $\Pi(t)$ be the average probability that a super constraint node sends a failure message, 
i.e., that it can not correct external errors lying in its domain. Then, the probability that a noisy pattern neuron 
with degree $d_i$ sends an erroneous message to a particular neighbor among super constraint node is equal to the 
probability that none of its \emph{other} neighboring super constraint nodes could have corrected its error, i.e.,
\begin{displaymath}
P_i(t) = p_e (\Pi(t))^{d_i-1}.
\end{displaymath}
Averaging over $d_i$ we find the average probability of error in iteration $t$:
\begin{equation}\label{eq:prob_error_z}
z(t+1) = p_e \widetilde{\lambda}(\Pi(t)) \mbox{.}
\end{equation}

Now consider a cluster $\ell$ that contains $d_\ell$ pattern neurons. This cluster will \emph{not} send a failure 
message over its edge to a noisy pattern neuron in its domain with probability:
\begin{enumerate}
\item $P_{c_1}$, if it is not connected to any other noisy neuron;
\item $P_{c_2}$, if it is connected to exactly one other constraint neuron;
\item $P_{c_3}$, if it is connected to exactly two other constraint neurons; and
\item $0$, if it is connected to more than two other constraint neuron.
\end{enumerate}
Thus,
\[
\Pi^{(\ell)}(t) = 1-P_{c_1}\left(1-z(t)\right)^{d_\ell-1} -P_{c_2}{d_\ell-1 \choose 1} z(t)\left(1-z(t)\right)^{d_\ell-2}-P_{c_3}{d_\ell-1 \choose 2}z(t)^2\left(1-z(t)\right)^{d_\ell-3}\mbox{.}
\]
Averaging over $d_\ell$ yields:
\begin{equation}
\label{eq:prob_error_Pi}
\Pi(t) =  \mathbb{E}_{d_\ell}\left(\Pi^{(\ell)}(t) \right) = 1 - P_{c_1} \rho(1-z(t)) -P_{c_2} z\rho^{\prime}(1-z(t)) -\tfrac{1}{2} P_{c_2} z(t)^2\rho^{\prime\prime}\left(1-z(t) \right)\mbox{,}
\end{equation}
where $\rho^{\prime}(x)$ and $\rho^{\prime\prime}(x)$ are derivatives of the function $\rho(x)$ with respect to $x$.

Equations \eqref{eq:prob_error_z} and \eqref{eq:prob_error_Pi} yield the value of $z(t+1)$ as a function of $z(t)$. 
We calculate the final error probability as $\lim_{t\to\infty} z(t)$; for $\lim_{t\to\infty} z(t) \rightarrow 0$, 
it is sufficient to have $z(t+1) < z(t)$, which proves the theorem.
\end{IEEEproof}
It must be mentioned that the above theorem holds when the decision subgraphs for the pattern neurons in graph 
$\widetilde{G}$ are tree-like for a depth of $\tau L$, where $\tau$ is the total number of number of iterations 
performed by Algorithm \ref{algo:peeling} \cite{RichardsonU2008}.

Theorem~\ref{th:outside} states that for any fraction of errors $\Lambda_{\upsilon,\nu}\leq \LnI$ that satisfies the above 
recursive formula, Algorithm~\ref{algo:peeling} will be successful with probability close to one. Note that the first fixed 
point of the above recursive equation dictates the maximum fraction of errors $\LnI$ that our model can correct. For the 
special case of $P_{c_1} = 1$ and $P_{c_i}=0$, for all $i > 1$, we obtain $\epsilon \widetilde{\lambda}\left(1-\widetilde{\rho}(1-z)\right) < z$, 
the same condition given in \cite{KarbasiSS2013}. Theorem~\ref{th:outside} takes into account the contribution of all $P_{c_i}$ 
terms and as we will see, their values change as we incorporate the effect of internal noise $\upsilon$ and $\nu$. 
Our results show that the maximum value of $P_{c_i}$ does not occur when the internal noise is equal to zero, i.e.\ 
$\upsilon=\nu = 0$, but instead when the neurons are contaminated with internal noise! As an example, 
Figure~\ref{fig:P_c_i_calculation} illustrates how $P_{c_i}$ behaves as a function of $\upsilon$ in the network considered 
(note that maximum values are not at $\upsilon=0$). This finding suggests that even individual clusters are able to correct 
more errors in the presence of internal noise.

To estimate the $P_{c_i}$ values, we use numerical approaches.\footnote{Appendix~\ref{sec:P_c_1_estimate} derives an 
analytical upper bound to estimate $P_{c_1}$ but this requires approximations that are loose.}
Given a set of clusters $W^{(1)},\dots,W^{(L)}$, for each cluster we randomly corrupt $i$ pattern neurons with 
$\pm 1$ noise. Then, we apply Algorithm~\ref{algo:correction} over this cluster and calculate the success rate once finished. 
We take the average of this rate over all clusters to end up with $P_{c_i}$. The results of this approach are shown in 
Figure~\ref{fig:P_c_i_calculation}, where the value of $P_{c_i}$ is shown for $i=1,\dots,4$ and various noise amounts at 
the pattern neurons (specified by parameter $\upsilon$). 

\subsection{Simulations}
\label{sec:simulations} 
Now we consider simulation results for a finite system.  To learn the subspace constraints \eqref{eq:orthogonal} 
for each cluster $G^{(\ell)}$ we use the learning algorithm in \cite{KarbasiSS2013}. Henceforth, we assume that 
the weight matrix $W$ is known and given. In our setup, we consider a network of size $n=400$ with $L=50$ clusters. 
We have $40$ pattern nodes and $20$ constraint nodes in each cluster, on average. External error is modeled by 
randomly generated vectors $z$ with entries $\pm 1$ with probability $\epsilon$ and $0$ otherwise. Vector $z$ is 
added to the correct patterns, which satisfy \eqref{eq:orthogonal}. For recall, Algorithm~\ref{algo:peeling} is used 
and results are reported in terms of Symbol Error Rate (SER) as the level of external error ($\epsilon$) or 
internal noise ($\upsilon, \nu$) is changed; this involves counting positions where the output of Algorithm~\ref{algo:peeling} 
differs from the correct pattern. 

\subsubsection{Symbol Error Rate as a function of Internal Noise}

\begin{figure}
  \centering
  \includegraphics[width=5in]{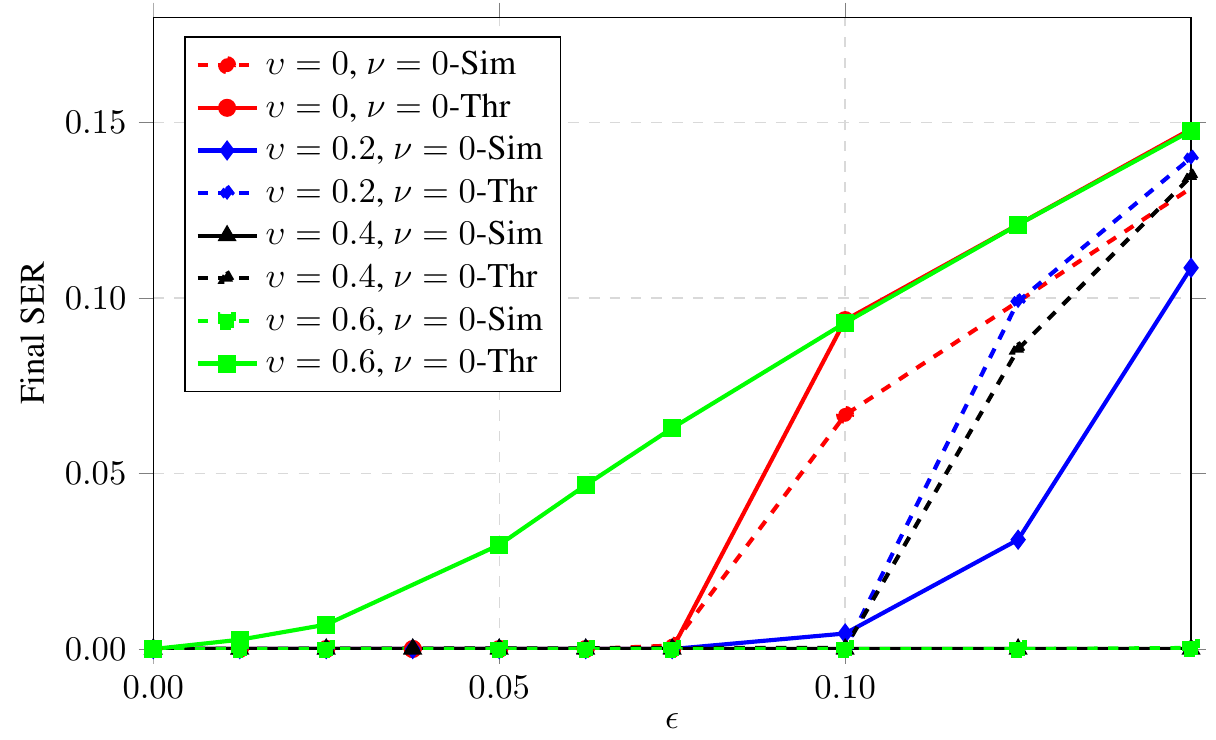}
  \caption{The final SER for a network with $n=400$, $L=50$. The red curves correspond to the noiseless neural network.}
  \label{fig:recall_error_rate}
\end{figure}

Figure~\ref{fig:recall_error_rate} illustrates the final SER of our algorithm for different values of 
$\upsilon$ and $\nu$. Remember that $\upsilon$ and $\nu$ quantify the level of noise in pattern and 
constraint neurons, respectively. Dashed lines in Figure~\ref{fig:recall_error_rate} are simulation results 
whereas solid lines are theoretical upper bounds provided in this paper. As evident, there is a threshold 
phenomenon such that SER is negligible for $\epsilon \leq \epsilon^*$ and grows beyond this threshold. 
As expected, simulation results are better than the theoretical bounds. In particular, the gap is relatively 
large as $\upsilon$ moves towards one. 

A more interesting trend in Figure~\ref{fig:recall_error_rate} is the fact that internal noise helps in 
achieving better performance, as predicted by theoretical analysis (Theorem~\ref{th:noisy_decoder_as_good_as}). 
Notice how $\epsilon^*$ moves towards one as $\nu$ increases.

\begin{figure}
  \centering
  \includegraphics[width=5in]{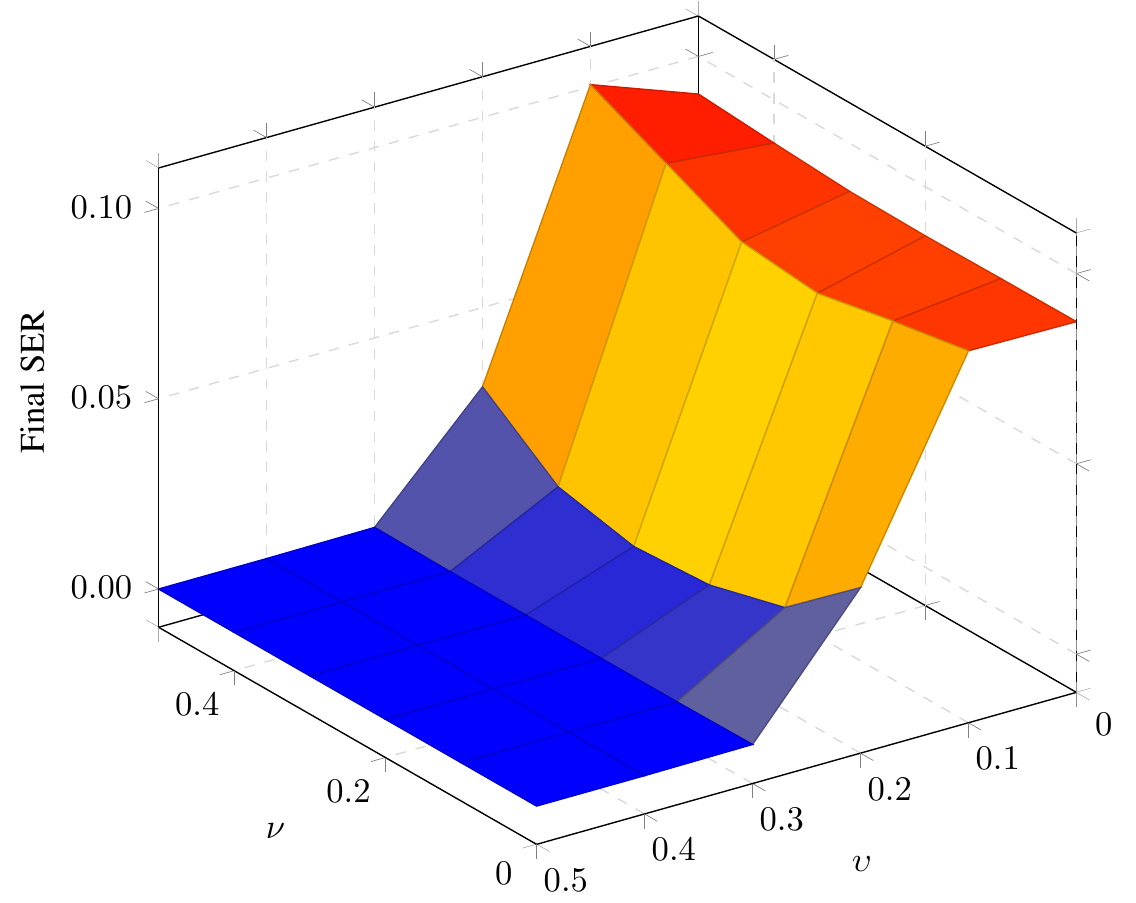}
  \caption{The final bit error probability when $\epsilon = 0.125$ as a function of internal noise parameters at the pattern and constraint neurons, denoted by $\upsilon$ and $\nu$, respectively.}
  \label{fig:3Dplot_BER_epsilon_0_125}
\end{figure}

\begin{figure}
  \centering
  \begin{subfigure}[b]{0.6\textwidth}
    \includegraphics[width=0.99\textwidth]{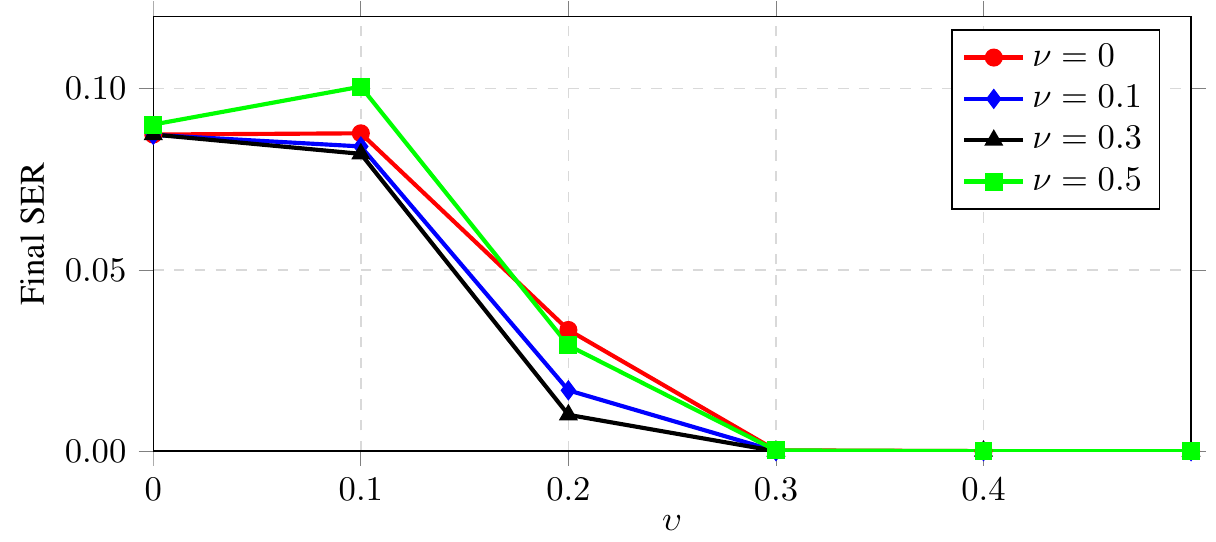}
    \caption{Final SER as function of $\upsilon$ for $\epsilon = 0.125$.}
    \label{fig:error_rate_epsilone_0_125_effect_of_upsilon}
  \end{subfigure}

\

  \begin{subfigure}[b]{0.6\textwidth}
    \includegraphics[width=0.99\textwidth]{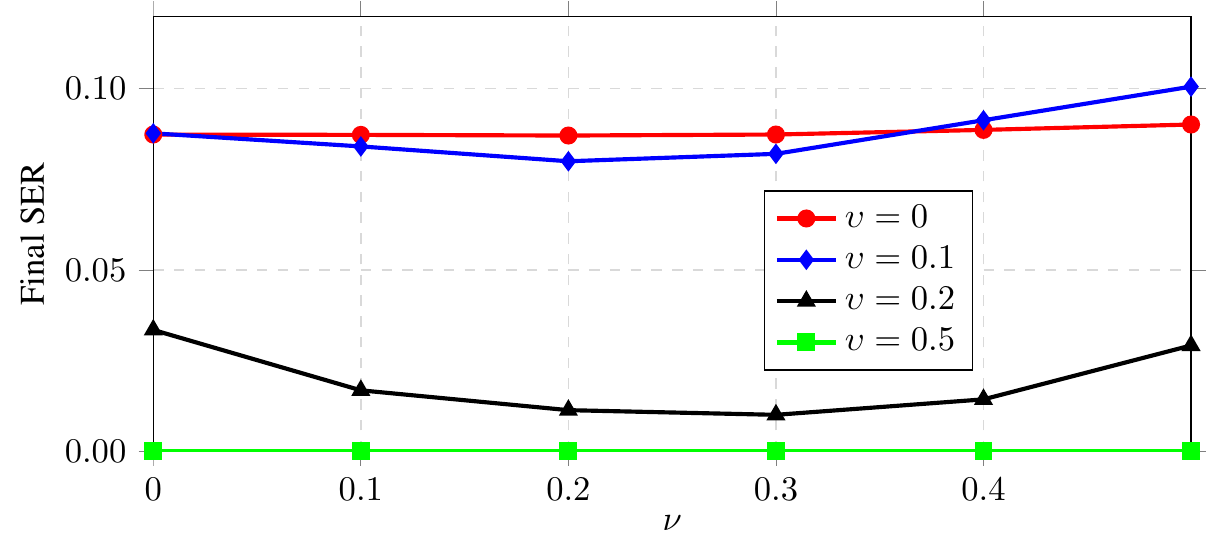}
    \caption{The effect of $\nu$ on the final SER for $\epsilon = 0.125$}        
    \label{fig:error_rate_epsilone_0_125_effect_of_nu}
  \end{subfigure}
  \caption{The final bit error probability as a function of internal noise parameters at pattern and constraint neurons for $\epsilon = 0.125$}
  \label{fig:error_rate_epsilone_0_125}
\end{figure}

This phenomenon is inspected more closely in Figure~\ref{fig:3Dplot_BER_epsilon_0_125} where $\epsilon$ is fixed to 
$0.125$ while $\upsilon$ and $\nu$ vary. Figs.~\ref{fig:error_rate_epsilone_0_125_effect_of_upsilon} and 
\ref{fig:error_rate_epsilone_0_125_effect_of_nu} display projected versions of the surface plot to investigate the 
effect of $\upsilon$ and $\nu$ separately. As we see again, a moderate amount of internal noise at both pattern 
and constraint neurons improves performance. There is an optimum point $(\upsilon^*, \nu^*)$ for which the SER 
reaches its minimum. Figure~\ref{fig:error_rate_epsilone_0_125_effect_of_nu} indicates for instance that $\nu^*\approx 0.25$, 
beyond which SER deteriorates. There is greater sensitivity to noise $\upsilon$ in the pattern neurons,
reminiscent of results for decoding circuits with internal noise \cite{TabatabaeiYazdiCD2013}.

\subsubsection{Larger noise values}
\label{sec:larger}

\begin{figure}
  \centering
  \includegraphics[width=5in]{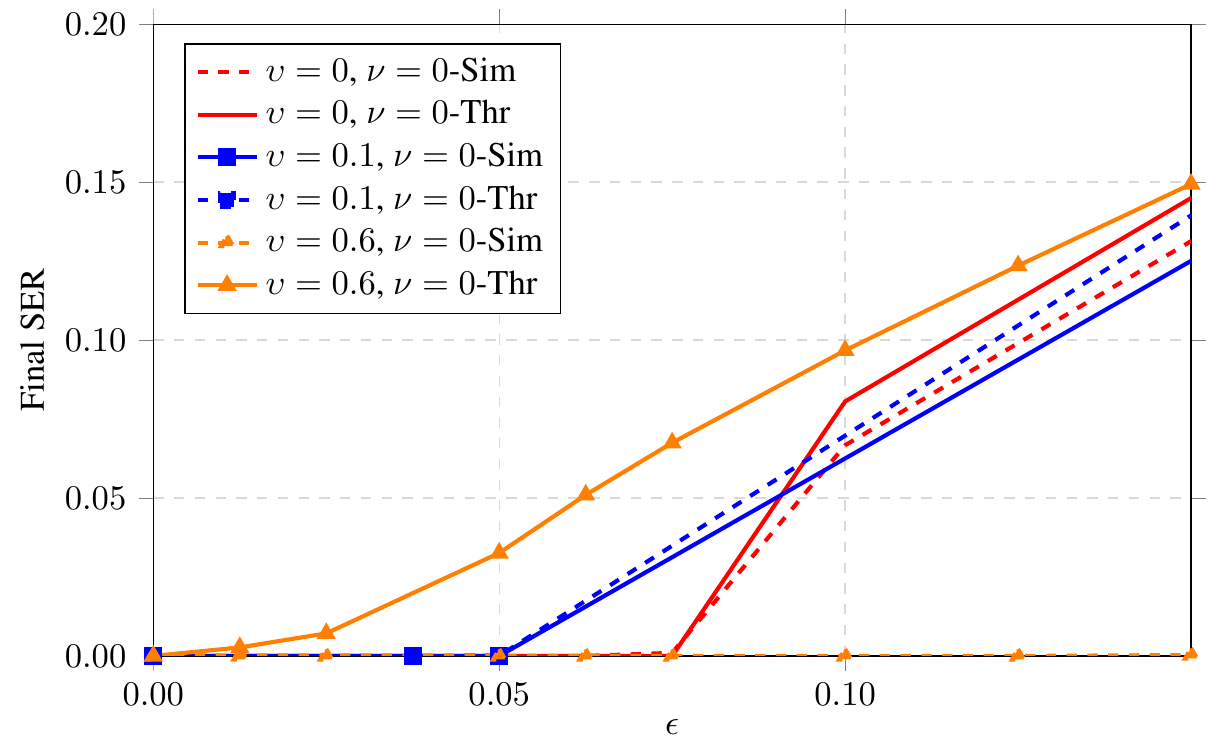}
  \caption{The final SER for a network with $n=400$, $L=50$ and noise values chosen from $\{-3,-2,\dots,2,3\}$. The blue curves correspond to the noiseless neural network.}
  \label{fig:noisy_neural_SER_larger_noise}
\end{figure}

So far, we have investigated the performance of the recall algorithm when noise values are limited to $\pm 1$. 
Although this choice facilitates the analysis of the algorithm and increases error correction speed, our analysis 
is valid for larger noise values. Figure~\ref{fig:noisy_neural_SER_larger_noise} illustrates the SER for the same 
scenario as before but with noise values chosen from $\{-3,-2,\ldots,2,3\}$. We see exactly the same behavior 
as we witnessed for $\pm 1$ noise values. 

\subsection{Recall Time as a function of Internal Noise}

\begin{figure}
  \centering
  \includegraphics[width=5in]{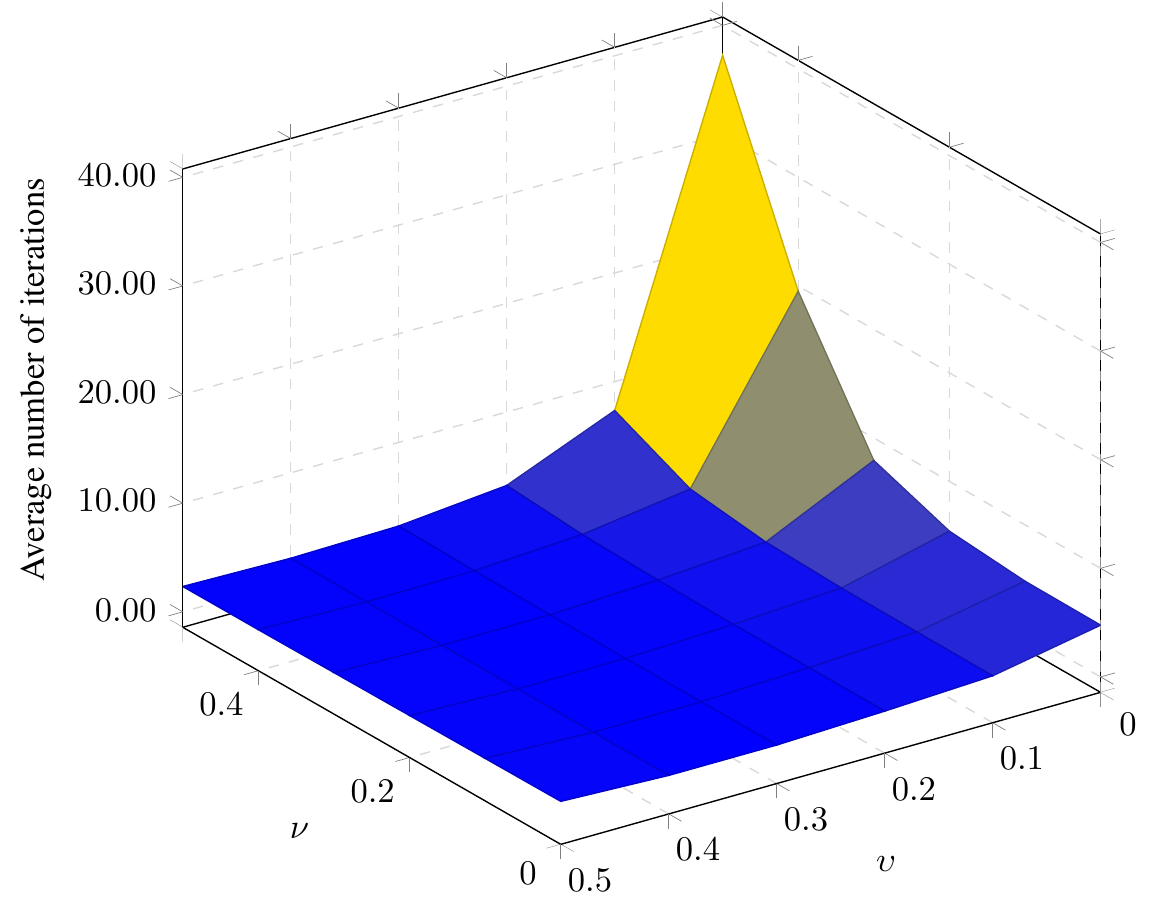}
  \caption{The effect of internal noise on the number of iterations performed by Algorithm~\ref{algo:peeling}, for different values of $\upsilon$ and $\nu$ with $\epsilon = 0.075$.}
  \label{fig:3Dplot_global_itr_epsilon_0_75}
\end{figure}

\begin{figure}
  \centering
  \begin{subfigure}[b]{0.6\textwidth}
    \includegraphics[width=0.99\textwidth]{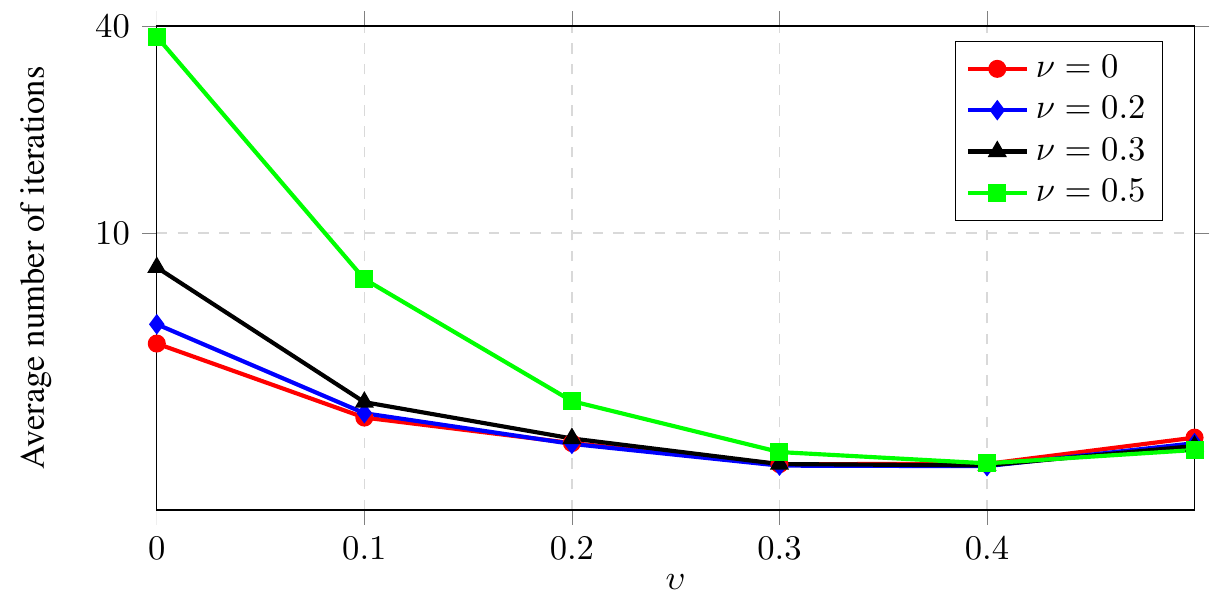}
    \caption{Effect of internal noise at pattern neurons side.}        
    \label{fig:Global_Itr_vs_upsilon_epsilon_0_75}
  \end{subfigure}

\

  \begin{subfigure}[b]{0.6\textwidth}
    \includegraphics[width=0.99\textwidth]{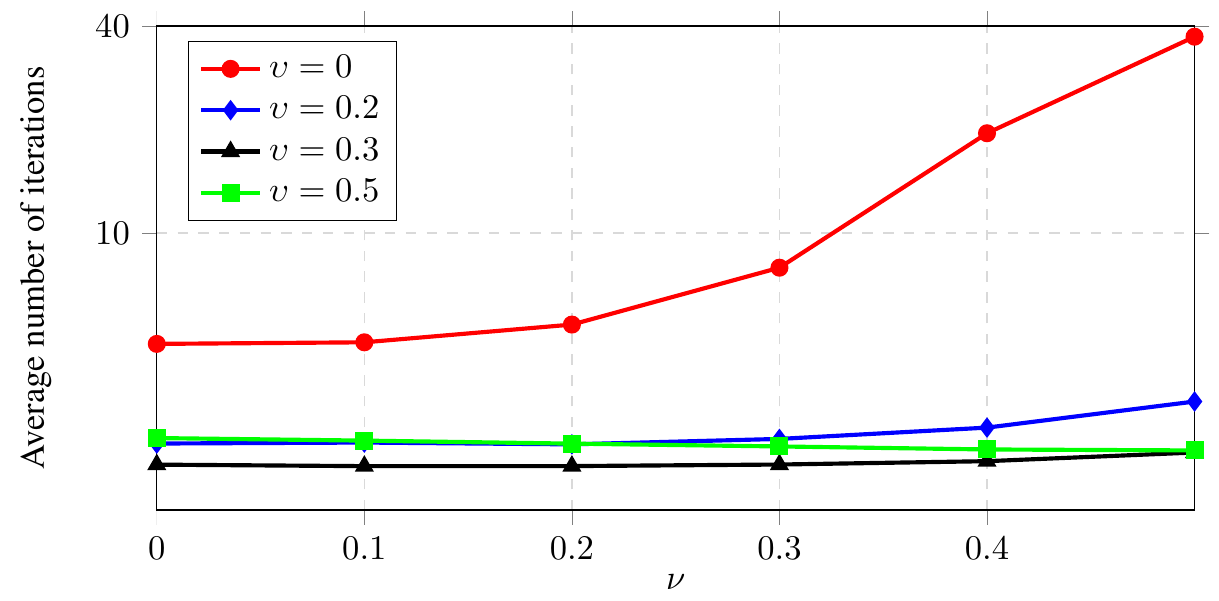}
    \caption{Effect of internal noise at constraint neurons side.}        
    \label{fig:Global_Itr_vs_nu_epsilon_0_75}
  \end{subfigure}
  \caption{The effect of internal noise on the number of iterations performed by Algorithm~\ref{algo:peeling}, for different values of $\upsilon$ and $\nu$ with $\epsilon =0.075$. The average iteration number of $40$ indicate the failure of Algorithm~\ref{algo:peeling}.}
  \label{Global_Itr_for_epsilon_0_75}
\end{figure}

Figure~\ref{fig:3Dplot_global_itr_epsilon_0_75} illustrates the number of iterations performed by Algorithm~\ref{algo:peeling} 
for correcting the external errors when $\epsilon$ is fixed to $0.075$.  We stop whenever the algorithm corrects all 
external errors or declare a recall error if all errors were not corrected in $40$ iterations. Thus, the corresponding 
areas in the figure where the number of iterations reaches $40$ indicates decoding failure. 
Figs.~\ref{fig:Global_Itr_vs_upsilon_epsilon_0_75} and \ref{fig:Global_Itr_vs_nu_epsilon_0_75} are projected versions of 
Figure~\ref{fig:3Dplot_global_itr_epsilon_0_75} and show the average number of iterations as a function of $\upsilon$ 
and $\nu$, respectively.

The amount of internal noise drastically affects the speed of Algorithm~\ref{algo:peeling}. First, from 
Figure~\ref{fig:3Dplot_global_itr_epsilon_0_75} and \ref{fig:Global_Itr_vs_nu_epsilon_0_75} observe that running time 
is more sensitive to noise at constraint neurons than pattern neurons and that the algorithms become slower as noise 
at constraint neurons is increased. In contrast, note that internal noise at the pattern neurons may improve the 
running time, as seen in Figure~\ref{fig:Global_Itr_vs_upsilon_epsilon_0_75}.  Ordering of sensitivity to noise in pattern neurons 
and in constraint neurons is opposite for running time as compared to error probability.

\begin{figure}
  \centering
  \includegraphics[width=5in]{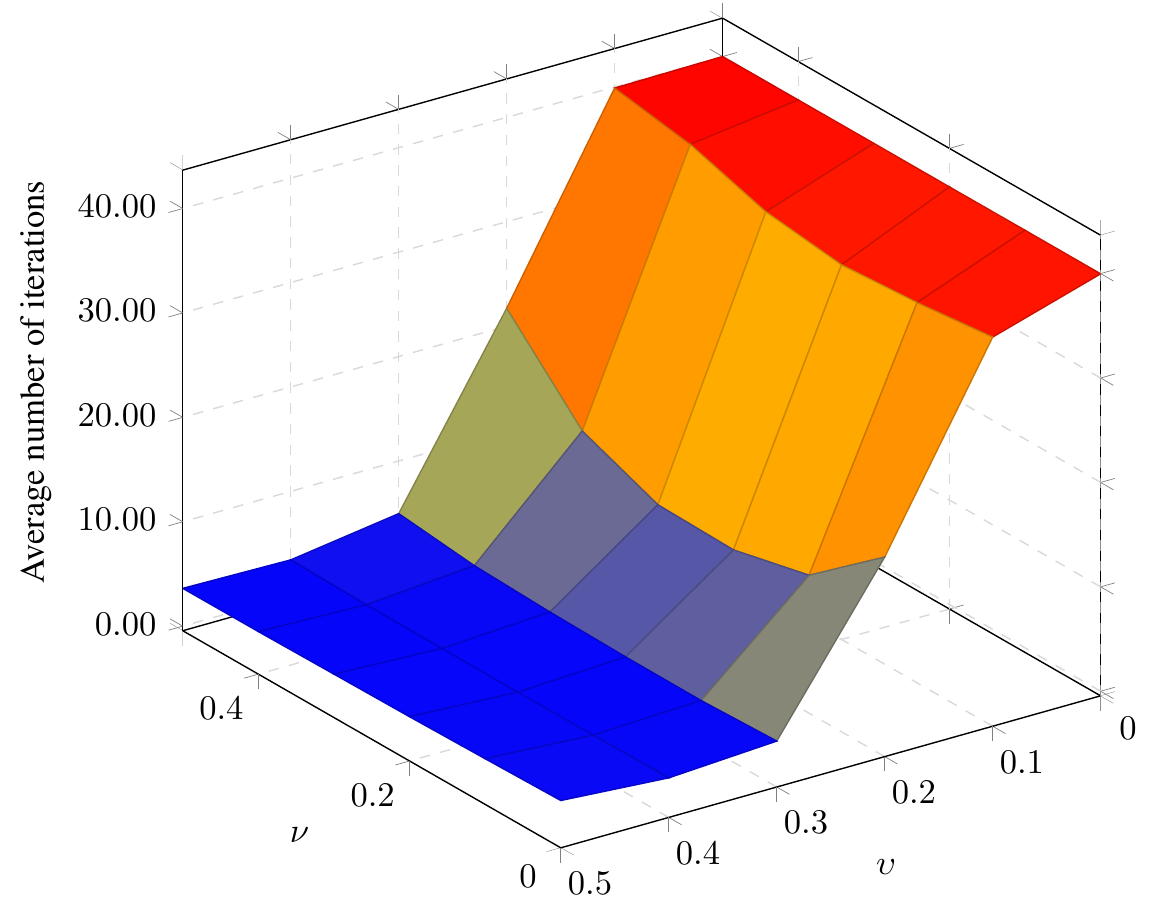}
  \caption{The effect of internal noise on the number of iterations performed by Algorithm~\ref{algo:peeling}, for different values of $\upsilon$ and $\nu$ with $\epsilon = 0.125$.}
  \label{fig:3Dplot_global_itr_epsilon_0_125}
\end{figure}

\begin{figure}
  \centering
  \begin{subfigure}{0.6\textwidth}
    \includegraphics[width=0.99\textwidth]{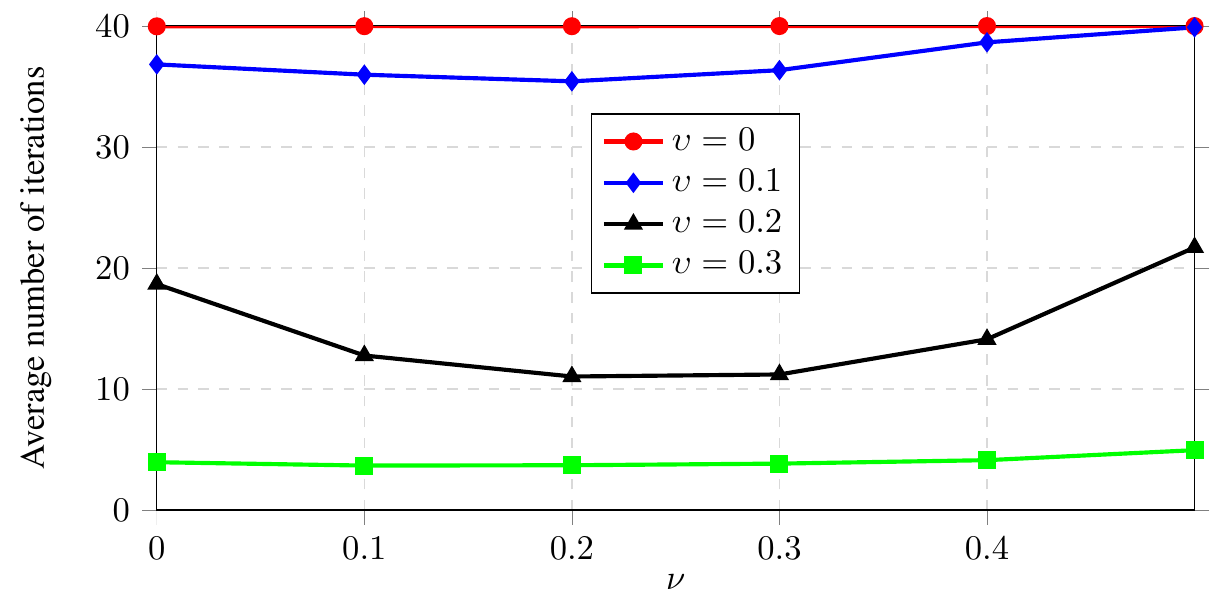}
    \caption{Effect of internal noise at constraint neurons side.}                
    \label{fig:Global_Itr_vs_nu_epsilon_0_125}
  \end{subfigure}

  \

  \begin{subfigure}{0.6\textwidth}
    \includegraphics[width=0.99\textwidth]{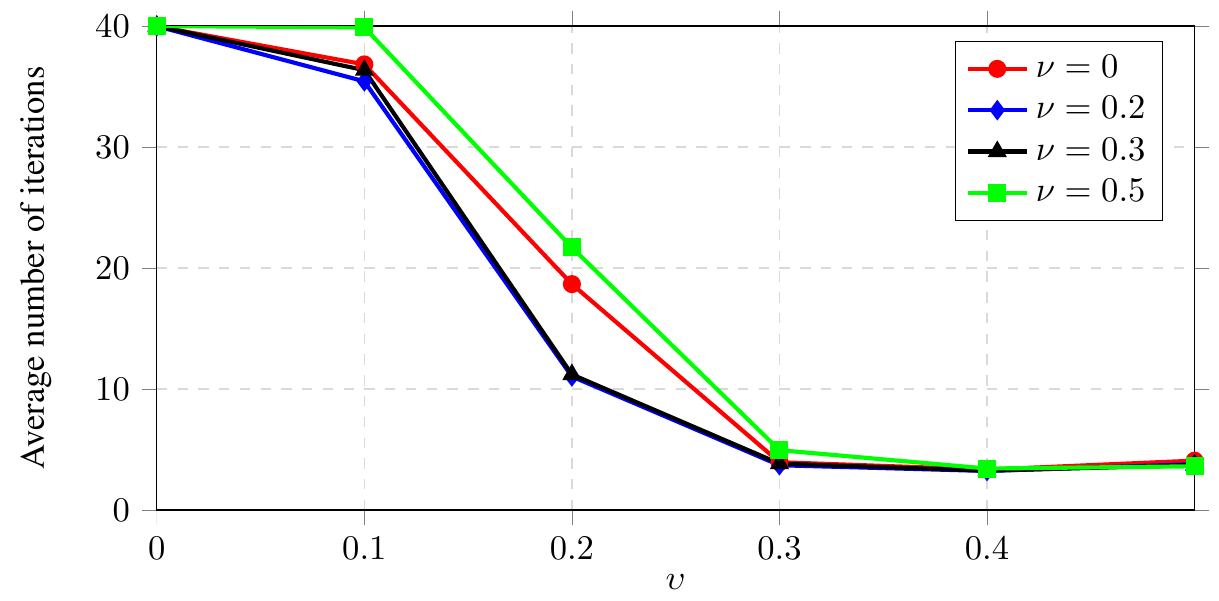}
    \caption{Effect of internal noise at pattern neurons side.}
    \label{fig:Global_Itr_vs_upsilon_epsilon_0_125}
  \end{subfigure}
  \caption{The effect of internal noise on the number of iterations performed by Algorithm~\ref{algo:peeling}, for different values of $\upsilon$ and $\nu$ with $\epsilon =0.125$. The average iteration number of $40$ indicate the failure of Algorithm~\ref{algo:peeling}.}
  \label{Global_Itr_for_epsilon_0_125}
\end{figure}

Note that the results presented so far are for the case where the noiseless decoder succeeds as well and 
its average number of iterations is pretty close to the optimal value (see Figure~\ref{fig:3Dplot_global_itr_epsilon_0_75}). 
Figure~\ref{fig:3Dplot_global_itr_epsilon_0_125} illustrates the number of iterations performed by 
Algorithm~\ref{algo:peeling} for correcting the external errors when $\epsilon$ is fixed to $0.125$. In this case, the 
noiseless decoder encounters stopping sets while the noisy decoder is still capable of correcting external errors. 
Here we see that the optimal running time occurs when the neurons have a fair amount of internal noise. Figs.~\ref{fig:Global_Itr_vs_upsilon_epsilon_0_125} and \ref{fig:Global_Itr_vs_nu_epsilon_0_125} are projected versions of 
Figure~\ref{fig:3Dplot_global_itr_epsilon_0_125} and show the average number of iterations as a function 
of $\upsilon$ and $\nu$, respectively.

\subsection{Effect of internal noise on the performance in absence of external noise}
Now we provide results of a study for a slightly modified setting where there is only internal noise and no 
external errors and further $\varphi < \upsilon$. Thus, the internal noise can now cause neurons to make 
wrong decisions, even in the absence of external errors. With abuse of notation, we assume pattern neurons are 
corrupted with a $\pm 1$ noise added to them with probability $\upsilon$. The rest of the model setting is the 
same as before. 

\begin{figure} 
  \centering
  \includegraphics[width=5in]{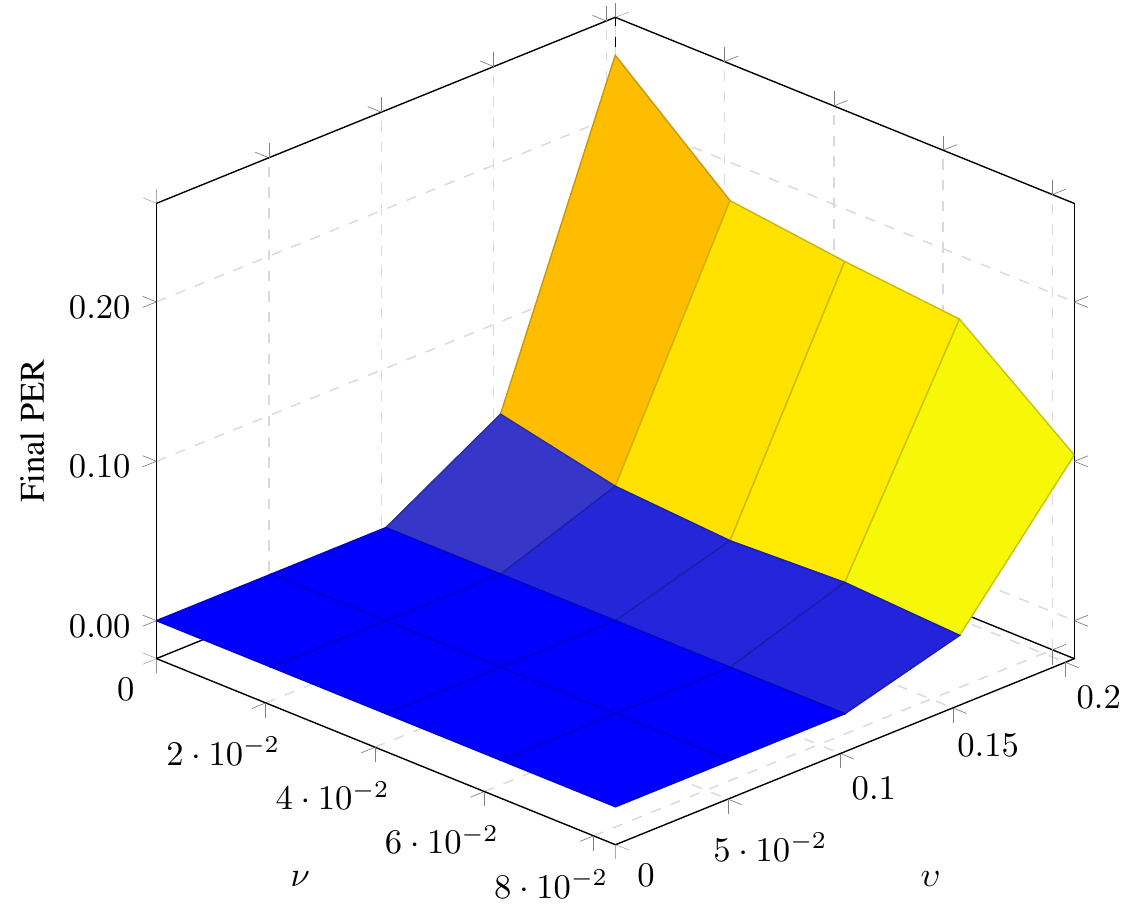}
  \caption{The effect of the internal noise on final Pattern Error Rate (PER) as a function of $\upsilon$ and $\nu$ in absence of external noise.}
  \label{fig:3Dplot_noiseless_case}
\end{figure}

\begin{figure}
  \centering
  \begin{subfigure}{0.6\textwidth}
    \includegraphics[width=0.99\textwidth]{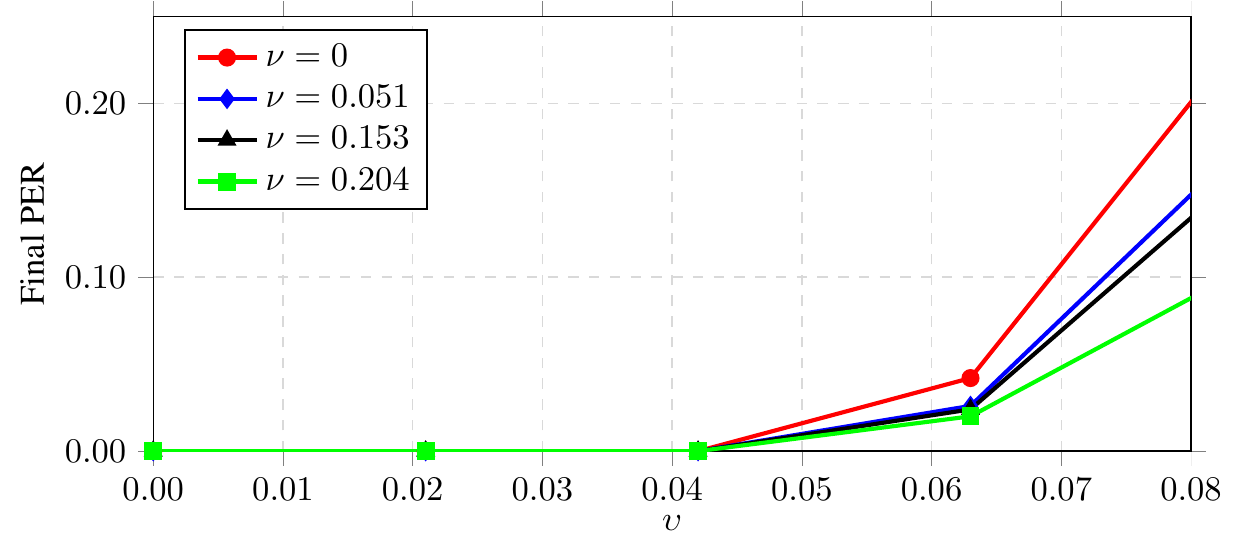}
    \caption{Effect of internal noise at pattern neurons side.}
    \label{fig:noiseless_case_effect_of_upsilon}
  \end{subfigure}

  \

  \begin{subfigure}{0.6\textwidth}
    \includegraphics[width=0.99\textwidth]{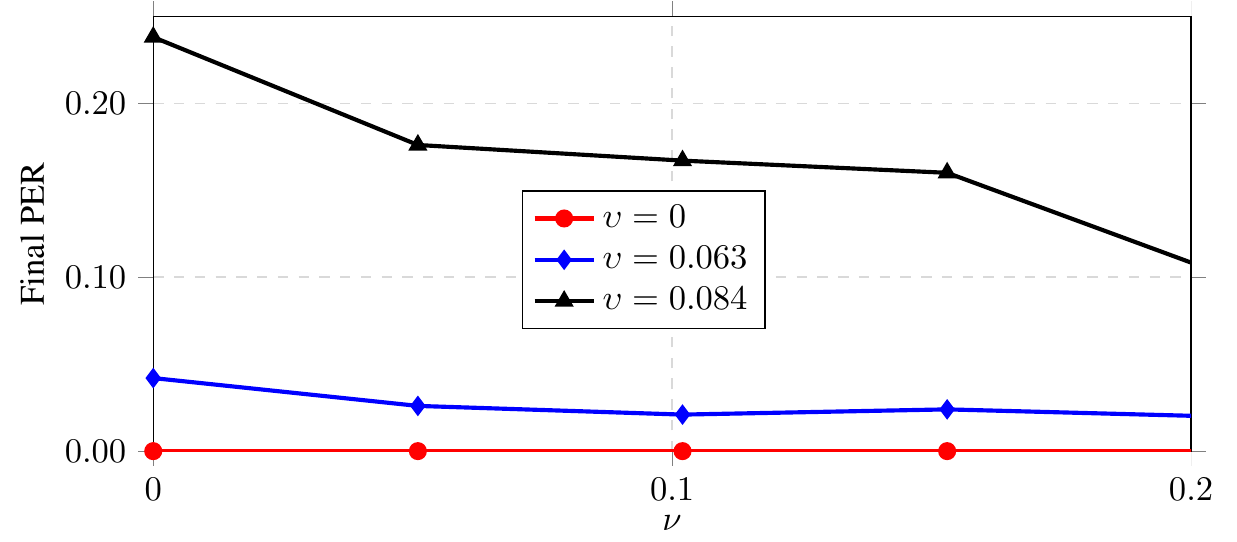}
    \caption{Effect of internal noise at constraint neurons side.}
    \label{fig:noiseless_case_effect_of_nu}
  \end{subfigure}
  \caption{The effect of the internal noise on final Pattern Error Rate (PER) as a function of $\upsilon$ and $\nu$ in absence of external noise.}
  \label{fig:noiseless_case_effect_of_upsilon_and_nu}
\end{figure}

Figure~\ref{fig:3Dplot_noiseless_case} illustrates the effect of internal noise as a function of $\upsilon$ and $\nu$, 
the noise parameters at the pattern and constraint nodes, respectively. This behavior is shown in 
Figs.~\ref{fig:noiseless_case_effect_of_upsilon} and \ref{fig:noiseless_case_effect_of_nu} for better inspection.
Here, we witness the more familiar phenomenon where increasing the amount of internal noise results in a worse 
performance. This finding emphasizes the importance of choosing update threshold $\varphi$ and $\psi$ properly, according 
to Lemma~\ref{lem:pi_0}.  See Appendix~\ref{section:choosing_gamma_appendix} for details on choosing thresholds.

\section{Discussion}
\label{sec:conclusions} 

We have demonstrated that associative memories with exponential capacity still work reliably even when built 
from unreliable hardware, addressing a major problem in fault-tolerant computing and further arguing for the 
viability of associative memory models for the (noisy) mammalian brain.  After all, brain regions modeled as 
associative memories, such as the hippocampus and the olfactory cortex, certainly do display internal noise 
\cite{Koch1999,YoshidaHTI2002,McDonnellW2011}.  

The linear-nonlinear computations of Algorithm~\ref{algo:correction} are nearly identical to message-passing algorithms such
as belief propagation and are certainly biologically plausible \cite{BeckP2007, DayanHNZ1995, Deneve2008, 
DoyaIPR2007, HintonS1986, MaBLP2006, LitvakU2009}.  The state reversion computation of Algorithm~\ref{algo:peeling} requires
keeping a state variable for a short amount of time which has been suggested as realistic for biological neurons
\cite{DruckmannC2012}, but the general biological plausibility of Algorithm~\ref{algo:peeling} remains an open question. 

We found a threshold phenomenon for reliable operation, which manifests the tradeoff between the amount of 
internal noise and the amount of external noise that the system can handle.  In fact, we showed that internal noise 
actually improves the performance of the network in dealing with external errors, up to some optimal value.  This 
is a manifestation of the \emph{stochastic facilitation} \cite{McDonnellW2011} or \emph{noise enhancement} 
\cite{ChenVKM2007} phenomenon that has been observed in other neuronal and signal processing systems, providing 
a functional benefit to variability in the operation of neural systems.  

The associative memory design developed herein uses thresholding operations in the message-passing algorithm for recall; 
as part of our investigation, we optimized these neural firing thresholds based on the statistics of the internal noise. 
As noted by Sarpeshkar in describing the properties of analog and digital computing circuits, ``In a cascade of analog stages, 
noise starts to accumulate. Thus, complex systems with many stages are difficult to build. [In digital systems] Round-off error 
does not accumulate significantly for many computations. Thus, complex systems with many stages are easy to build'' \cite{Sarpeshkar1998}. 
One key to our result is capturing this benefit of digital processing (thresholding to prevent the build up of errors due to 
internal noise) as well as a modular architecture which allows us to correct a linear number of external errors (in terms 
of the pattern length).

This paper focused on recall, however learning is the other critical stage of associative memory operation.
Indeed, information storage in nervous systems is said to be subject to storage (or learning) noise, \emph{in situ} noise, 
and retrieval (or recall) noise \cite[Figure~1]{VarshneySC2006}. It should be noted, however, there is no essential 
loss by combining learning noise and \emph{in situ} noise into what we have called external error herein, 
cf.~\cite[Fn.~1 and Prop.~1]{Varshney2011}. Thus our basic qualitative result extends to the setting 
where the learning and stored phases are also performed with noisy hardware.

Going forward, it is of interest to investigate other neural information processing models that explicitly incorporate
internal noise and see whether they provide insight into observed empirical phenomena.  As an example, we might be able 
to explain the threshold phenomenon observed in the symbol error rate of human telegraph operators under heat stress 
\cite[Figure~2]{Mackworth1946}, by invoking a thermal internal noise explanation.
Returning to engineering, internal noise in decoders for limited-length error-correcting codes may improve performance 
as observed herein, since stopping sets are a limiting phenomenon in that setting also.

\section*{Acknowledgment}
We thank S.~S.\ Venkatesh for telling us about \cite{Biswas1993}.

\appendices

\section{Illustrating Proof of Theorem \ref{th:noisy_decoder_as_good_as}}
\label{sec:proof_thm_noisy_decoder_as_good_as}

\begin{figure}
  \centering
  \begin{subfigure}[b]{0.48\textwidth}
    \includegraphics[width=.95\textwidth]{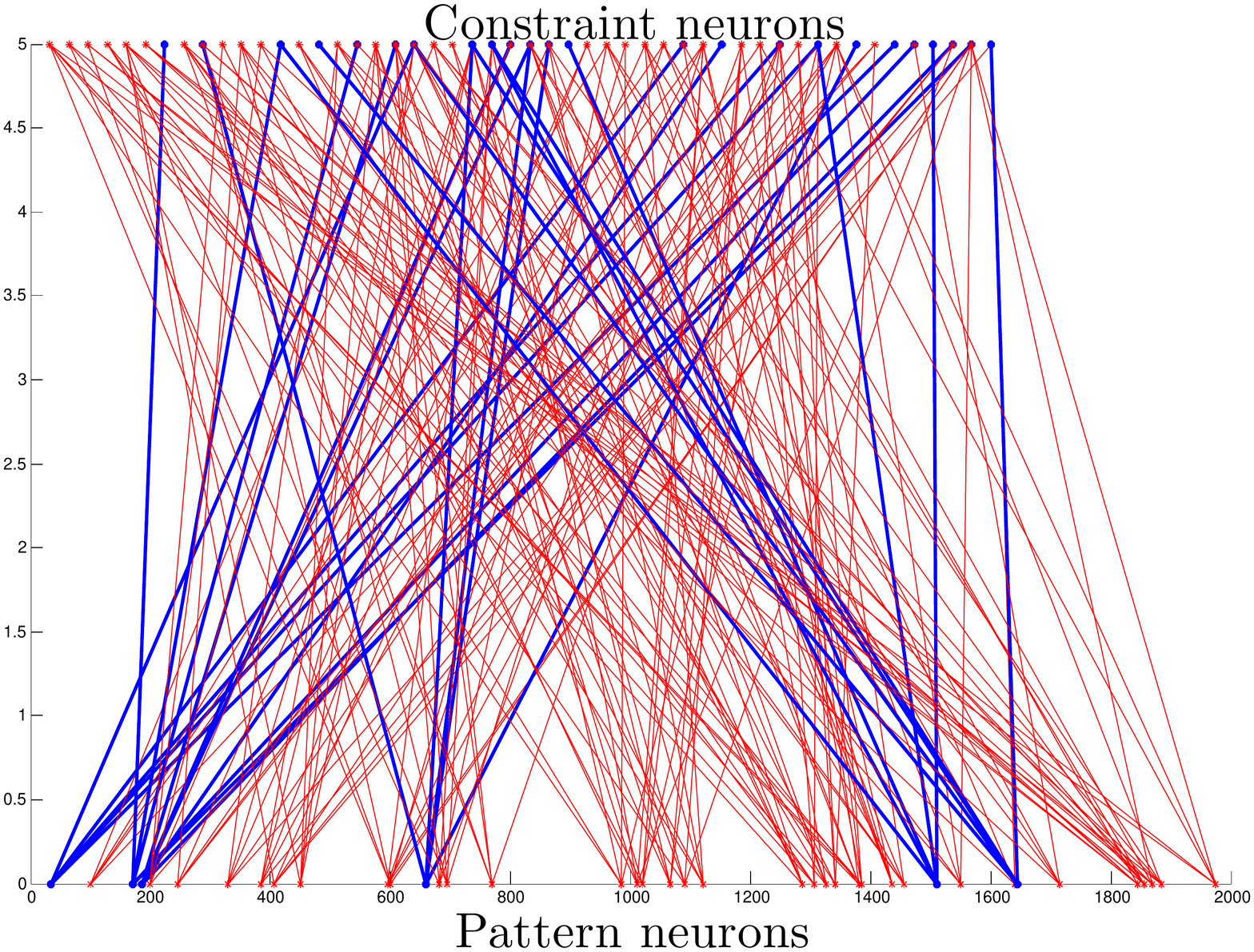}
    \caption{$t = 0$}
    \label{fig:stopping_set_itr_0}
  \end{subfigure}~
  \begin{subfigure}[b]{0.48\textwidth}
    \includegraphics[width=.95\textwidth]{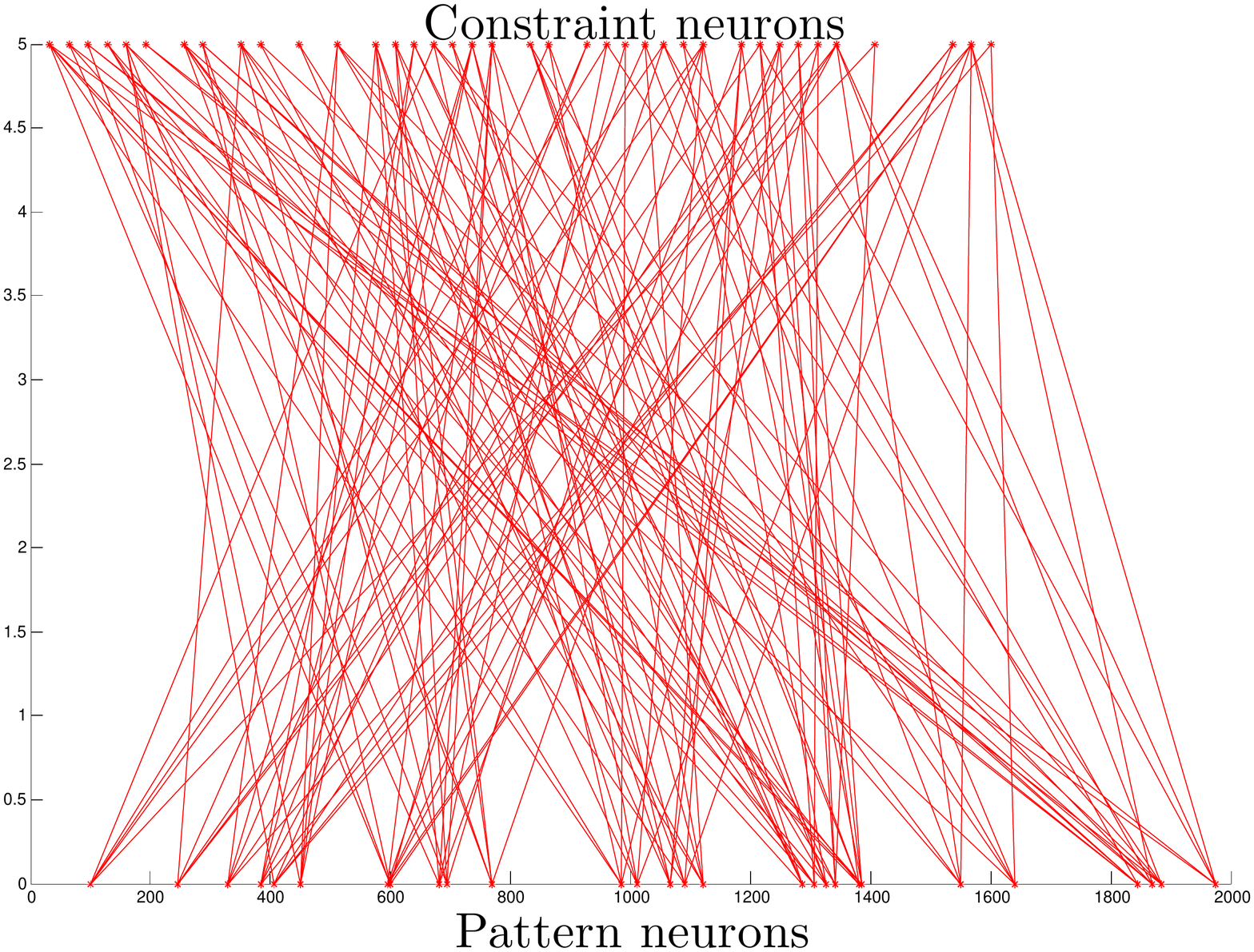}
    \caption{$t \rightarrow \infty$}
    \label{fig:stopping_set_itr_infty}
  \end{subfigure}
  \caption{An external noise pattern that contains a stopping set in a noiseless neural circuit. Left figure shows the original pattern and the right figure illustrates the result of the decoding algorithm after sufficient number of iterations where the algorithm gets stuck. Blue pattern nodes are those that are connected to at least one cluster with a single external error. Obviously, the stopping set on the right does not have any blue nodes.}
  \label{fig:stopping_set}
\end{figure}

Figure~\ref{fig:stopping_set_itr_0} illustrates an example of a stopping set over the graph $\widetilde{G}$ in our empirical 
studies. In the figure, only the nodes corrupted with external noise are shown for clarity. Pattern neurons that are 
connected to at least one cluster with a single error are colored blue and other pattern neurons are colored red. 
Figure~\ref{fig:stopping_set_itr_infty} illustrates the same network but after a number of decoding iterations that result
in the algorithm getting stuck.  We have a stopping set in which no cluster has a single error and the algorithm cannot 
proceed further since $P_{c_i} \simeq 0$ for $i>1$ in a noiseless architecture. Thus, the external error cannot get corrected.

As evident from Figure~\ref{fig:stopping_set}, the stopping set is the result of clusters not being able to correct more 
than one external error; this is where internal noise might come to the rescue. Interestingly, an ``unreliable'' neural 
circuit in which $\upsilon = 0.6$ could easily get out of the stopping set shown in Figure~\ref{fig:stopping_set_itr_infty} 
and correct all of the external errors. Because we try several times to correct errors in a cluster 
(and overall in the network) while making sure that the algorithm does not introduce new errors itself. Thus, the noise 
might act in our favor in one of these attempts and the algorithm might be able to avoid stopping set as depicted in 
Figure~\ref{fig:stopping_set}.

\section{Estimating $P_{c_1}$ Theoretically}
\label{sec:P_c_1_estimate}

To bound $P_{c_1}$, consider four event probabilities for a cluster: 
\begin{itemize}
\item $\pi^{(\ell)}_0$ (resp. $P^{(\ell)}_0$): The probability that a constraint neuron (resp. pattern neuron) in 
cluster $\ell$ makes a wrong decision due to its internal noise when there is no external noise introduced to cluster 
$\ell$, i.e. $\Vert z^{(\ell)} \Vert_0 = 0$.
\item $\pi^{(\ell)}_1$ (resp. $P^{(\ell)}_1$): The probability that a constraint neuron (resp. pattern neuron) in 
cluster $\ell$ makes a wrong decision due to its internal noise when  one input error (external noise) is introduced, 
i.e.\ $\Vert z^{(\ell)} \Vert_0 = \Vert z^{(\ell)} \Vert_1 = 1$.
\end{itemize}
\noindent Notice $P^{(\ell)}_{c_1} = 1-P^{(\ell)}_1$.

We derive an upper bound on the probability a constraint node makes a mistake in the presence of one external error.
\begin{lemma}
\label{lem:pi_1}
In the presence of a single external error, the probability that a constraint neuron makes a wrong decision due to 
its internal noise is given by
\[
\pi^{(\ell)}_1 \leq \max \left(0,\frac{\nu -(\eta-\psi)}{2\nu} \right)\mbox{,}
\]
where $\eta = \min_{i,j,W_{ij}\neq 0} \left(\vert W_{ij}\vert \right)$ is the minimum absolute value of the non-zero 
weights in the neural graph and is chosen such that $\eta \geq \psi$.\footnote{This condition can be enforced during 
simulations as long as $\psi$ is not too large, which itself is determined by the level of constraint neuron internal 
noise, $\nu$, as we must have $\psi \geq \nu$.}
\end{lemma}
\begin{IEEEproof}
Without loss of generality, assume it is the first pattern node, $x^{(\ell)}_1$, that is corrupted with noise $+1$. 
Now calculate the probability that a constraint node makes a mistake in such circumstances. We only need analyze 
constraint neurons connected to $x^{(\ell)}_1$ since the situation for other constraint neurons is as when there is
no external error.  For a constraint neuron $j$ connected to $x^{(\ell)}_1$, the decision parameter is
\begin{align*}
h^{(\ell)}_j &= \left(W^{(\ell)}.(x^{(\ell)}+z^{(\ell)})\right)_j + v_j \\ \notag 
&= 0 + \left( W^{(\ell)}.z^{(\ell)}\right)_j + v_j  \\ \notag
&=  w_{j1} + v_j \mbox{.} 
\end{align*}

We consider two error events:
\begin{itemize}
\item A constraint node $j$ makes a mistake and does not send a message at all. The probability of this event 
is denoted by $\pi_1^{(1)}$.
\item A constraint node $j$ makes a mistake and sends a message with the opposite sign. The probability of this event 
is denoted by $\pi_2^{(1)}$.
\end{itemize}

We first calculate the probability of $\pi_2^{(1)}$. Without loss of generality, assume the $w_{j1}>0$ so that the 
probability of an error of type two is as follows (the case for $w_{j1}<0$ is exactly the same):
\begin{align}
\label{Pi_2_1}
\pi_2^{(1)} &= \mbox{Pr}\{ w_{ji}+v_j < -\psi \} \\ \notag
&= \max \left(0,\frac{\nu - (\psi+w_{j1})}{2\nu} \right)\mbox{.}
\end{align}
However, since $\psi > \nu $ and $w_{j1}>0$, then $\nu - (\psi+w_{j1}) < 0$ and $\pi_2^{(1)} = 0$. Therefore, the 
constraint neurons will never send a message that has an opposite sign to what it should have. All that remains 
is to calculate the probability they remain silent by mistake.

To this end, we have
\begin{align}
\pi_1^{(1)} &= \mbox{Pr}\{ |w_{ji}+v_j| < \psi \} \\ \notag
&= \max \left(0,\frac{\nu + \min(\psi-w_{j1},\nu)}{2\nu} \right)\mbox{.}
\end{align}
This can be simplified if we assume that the absolute values of all weights in the network are bigger than a 
constant $\eta>\psi$. Then, the above equation will simplify to
\begin{equation}
\label{Pi_1_1}
\pi_1^{(1)} \leq \max \left(0,\frac{\nu -(\eta-\psi)}{2\nu} \right)\mbox{.}
\end{equation}

Putting the above equations together, we obtain:
\begin{equation}
\label{Pi_1}
\pi^{(1)} \leq \max \left(0,\frac{\nu -(\eta-\psi)}{2\nu} \right)\mbox{.}
\end{equation}
\end{IEEEproof}

In the case $\eta-\psi > \nu$, we could even manage to make this probability equal to zero. However, we will 
leave it as is and use \eqref{Pi_1} to calculate $P_1^{(\ell)}$.

\subsection{Calculating $P^{(\ell)}_1$}
We start by calculating the probability that a non-corrupted pattern node $x^{(\ell)}_j$ makes a mistake, which 
is to change its state in round $1$. Let us denote this probability by $q^{(\ell)}_1$. Now to calculate $q^{(\ell)}_1$ 
assume $x^{(\ell)}_j$ has degree $d_j$ and it has $b$ common neighbors with $x^{(\ell)}_1$, the corrupted pattern node. 

Out of these $b$ common neighbors, $b_c$ will send $\pm 1$ messages and the others will, mistakenly, send nothing. Thus, 
the decision making parameter of pattern node $j$, $g^{(\ell)}_j$, will be bounded by 
\[
g^{(\ell)}_j =  \frac{\left(\mbox{sign}(W^{(\ell)})^\top\cdot y^{(\ell)}\right)_j}{d_j} + u_j. \leq \frac{b_c}{d_j} + u_j\mbox{.}
\]
We denote $\left(\mbox{sign}(W^{(\ell)})^\top\cdot y^{(\ell)}\right)_j$ by $o_j$ for brevity from this point on.

In this circumstance, a mistake happens when $|g^{(\ell)}_j| \geq \varphi$. Thus
\begin{align}
\label{eq:P_1_1_step_1}
q^{(\ell)}_1 &= \mbox{Pr}\{|g^{(\ell)}_j|\geq \varphi |\mbox{deg}(a_j) = d_j \& |\mathcal{N}(x_1) \cap \mathcal{N}(a_j) | = a\} \\ \notag
&= \mbox{Pr}\{\frac{o_j}{d_j} + u_j \geq \varphi \} + \mbox{Pr}\{\frac{o_j}{d_j} + u_j \leq - \varphi \} \mbox{,}
\end{align}
where $\mathcal{N}(a_i)$ represents the neighborhood of pattern node $a_i$ among constraint nodes. 

By simplifying \eqref{eq:P_1_1_step_1} we get
\[
q^{(\ell)}_1(o_j) = \begin{cases} +1, & \mbox{if } |o_j| \geq (\upsilon + \varphi) d_j
\\
\max(0,\frac{\upsilon-\varphi}{\upsilon}), & \mbox{if } |o_j| \leq |\upsilon - \varphi| d_j
\\
\frac{\upsilon - (\varphi - o_j/d_j)}{2\upsilon} , & \mbox{if } |o_j-\varphi d_j| \leq \upsilon d_j
\\
\frac{\upsilon - (\varphi + o_j/d_j)}{2\upsilon} , & \mbox{if } |o_j+\varphi d_j| \leq \upsilon d_j \mbox{.}
\end{cases}
\]

We now average this equation over $o_j$, $b_c$, $b$ and $d_j$. To start, suppose that out of the $b_c$ non-zero 
messages node $a_j$ receives, $e$ of them have the same sign as the link they are being transmitted over. 
Thus, we will have $o_j=e-(b_c-e) = 2e-b_c$. Assuming the probability of having the same sign for each message is 
$1/2$, the probability of having $e$ equal signs out of $b_c$ elements will be ${b_c\choose e} \left(\tfrac{1}{2}\right)^{b_c}$. 
Thus, we will get
\begin{equation}
\label{eq:P_1_1_step_3}
\bar{q}_{1}^{(\ell)}  = \sum_{e=0}^{b_c} {b_c\choose e} \left(\frac{1}{2}\right)^{b_c} q^{(\ell)}_1(2e-b_c)\mbox{.}
\end{equation}

Now note that the probability of having $a-b_c$ mistakes from the constraint side is given by 
${b \choose b_c} (\pi^{(\ell)}_1)^{b-b_c} (1-\pi^{(\ell)}_1)^{b_c}$. With some abuse of notations we get:
\begin{equation}
\label{eq:P_1_1_step_4}
\bar{q}_{1}^{(\ell)}  = \sum_{b_c =0}^{b} {b \choose b_c} (\pi^{(\ell)}_1)^{b-b_c} (1-\pi^{(\ell)}_1)^{b_c} \sum_{e=0}^{b_c} {b_c\choose e} \left(\frac{1}{2}\right)^{b_c} q^{(\ell)}_1(2e-b_c) \mbox{.}
\end{equation}

Finally, the probability that $a_j$ and $x_1$ have $b$ common neighbors can be approximated by 
${d_j\choose b} (1-\bar{d}^{(\ell)}/m_\ell)^{d_j-b} (\bar{d}^{(\ell)}/m_\ell)^{b}$, where 
$\bar{d}^{(\ell)}$ is the average degree of pattern nodes. Thus (again abusing some notation), we obtain:
\begin{equation}
\bar{q}_{1}^{(\ell)}  = \sum_{b=0}^{d_j} p_b \sum_{b_c =0}^{b} p_{b_c} \sum_{e=0}^{b_c} {b_c\choose e} \left(\frac{1}{2}\right)^{b_c} q^{(\ell)}_1(2e-b_c)\mbox{,}
\end{equation}
where $q^{(\ell)}_1(2e-b_c)$ is given by \eqref{eq:P_1_1_step_1}, $p_b$ is the probability of having $b$ 
common neighbors and is estimated by ${d_j\choose b} (1-\bar{d}^{(\ell)}/m_\ell)^{d_j-b} (\bar{d}^{(\ell)}/m_\ell)^{b}$, 
with $\bar{d}^{(\ell)}$ being the average degree of pattern nodes in cluster $\ell$. Furthermore, $p_{b_c}$ is the 
probability of having $b-b_c$ out of these $b$ nodes making mistakes. Hence, 
$p_{b_c}= {b \choose b_c} (\pi^{(\ell)}_1)^{b-b_c} (1-\pi^{(\ell)}_1)^{b_c}$. We will not simplify the above 
equation any further and use it as it is in our numerical analysis in order to obtain the best parameter $\varphi$.

Now we turn our attention to the probability that the corrupted node, $x_1$, makes a mistake, which is either 
not to update at all or update itself in the wrong direction. Recalling that we have assume the external noise 
term in $x_1$ to be a $+1$ noise, the wrong direction would be for node $x_1$ to increase its current value 
instead of decreasing it. Furthermore, we assume that out of $d_1$ neighbors of $x_1$, some $j$ of them have 
made a mistake and will not send any messages to $x_1$. Thus, the decision parameter of $x_1$, will be 
$g^{(\ell)}_1 = u + (d_1-j)/d_1$. Denoting the probability of making a mistake at $x_1$ by $q^{(\ell)}_2$ we get:
\begin{align}
\label{eq:P_1_2_step_1_original}
q^{(\ell)}_2 &= \mbox{Pr}\{g^{(\ell)}_1\leq \varphi | \mbox{deg}(x_1) = d_1 \mbox{ and } j\mbox{ errors in constraints} \} \\ \notag 
&= \mbox{Pr}\left\{\frac{d_1 -j}{d_1} + u < \varphi  \right\}\mbox{,}
\end{align}
which simplifies to
\begin{equation}
\label{eq:P_1_2_step_2_original}
q^{(\ell)}_2(j) = \begin{cases} +1, & \mbox{if } |j| \geq (1 + \upsilon - \varphi) d_1
\\
\max(0,\frac{\upsilon-\varphi}{\upsilon}), & \mbox{if } |j| \leq (1-\upsilon - \varphi) d_1
\\
\frac{\upsilon + \varphi - (d_1-j)/d_1}{2\upsilon} , & \mbox{if } |\varphi d_1-(d_1-j)| \leq \upsilon d_1\mbox{.}
\end{cases}
\end{equation}
Noting that the probability of making $j$ mistakes on the constraint side is 
${d_1\choose j} (\pi^{(\ell)}_1)^{j} (1-\pi^{(\ell)}_1)^{d_1-j}$, we get
\begin{equation}
\label{eq:P_1_2_step_3_original}
\bar{q}_{2}^{(\ell)} = \sum_{j=0}^{d_1} {d_1\choose j} (\pi^{(\ell)}_1)^{j} (1-\pi^{(\ell)}_1)^{d_1-j} q^{(\ell)}_2(j)\mbox{,}
\end{equation}
where $q^{(\ell)}_2(j)$ is given by \eqref{eq:P_1_2_step_2_original}.

Putting the above results together, the overall probability of making a mistake on the side of pattern neurons 
when we have one bit of external noise is
\begin{equation}
\label{P_e_1_original}
P^{(\ell)}_1 = \frac{1}{n^{(\ell)}} \bar{q}_{2}^{(\ell)} + \frac{n^{(\ell)}-1}{n^{(\ell)}} \bar{q}_{1}^{(\ell)} \mbox{.}
\end{equation}

Finally, the probability that cluster $\ell$ could correct one error is that all neurons take the correct decision, i.e.\
\[
P_{c_1}^{(\ell)} = (1-P^{(\ell)}_1)^{n^{(\ell)}}
\]
and the average probability that clusters could correct one error is simply
\begin{equation}
P_{c_1} = \mathbb{E}_\ell (P_{c_1}^{(\ell)}) \mbox{.}
\end{equation}

We use this equation in order to find the best update threshold $\varphi$.

\section{Choosing proper $\varphi$}
\label{section:choosing_gamma_appendix}
We now apply numerical methods to \eqref{P_e_1_original} to find the best $\varphi$ for different values of noise parameter $\upsilon$. The following figures show the best choice for the parameter $\varphi$. The update threshold on the constraint side is chosen such that $\psi > \nu$. In each figure, we have illustrated the final probability of making a mistake, $P_1^{(\ell)}$, for comparison.
\begin{figure}[t]
	\centering	
	\includegraphics[width=5in]{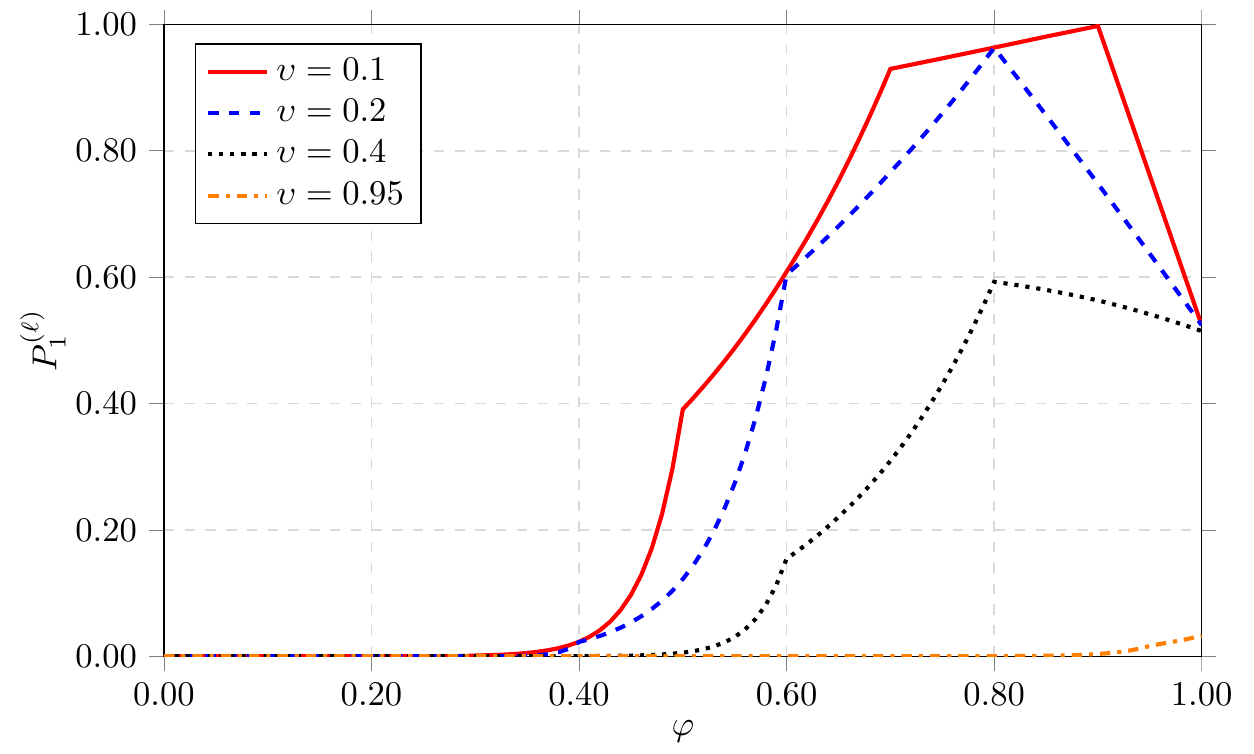}
	\centering
    	\caption{The behavior of $P_{c_1}$ as a function of $\varphi$ for different values of noise parameter, $\upsilon$. Here, $\pi_{(1)} = 0.01$.}    
\label{fig:error_prob_pi_1_01}
\end{figure}

Figure~\ref{fig:error_prob_pi_1_01} illustrates the behavior of the average probability of correcting a single error, $P_{c_1}$, as a function of $\varphi$ for different values of $\upsilon$ and for $\pi_1 = 0.01$. The interesting trend here is that in all cases, $\varphi^*$, the update threshold that gives the best result, is chosen such that it is quite large. This actually is in line with our expectation because a small $\varphi$ will result in non-corrupted nodes to update their states more frequently. On the other hand, a very large $\varphi$ will prevent the corrupted nodes to correct their states, especially if there are some mistakes made on the constraint side, i.e., $\pi^{(\ell)}_1>0$. Therefore, since we have much more non-corrupted nodes than corrupted nodes, it is best to choose a rather high $\varphi$ but not too high. Please also note that when $\pi^{(\ell)}_1$ is very high, there are no values of $\upsilon$ for which error-free storage is possible.

\begin{figure}
  \centering
  \includegraphics[width=5in]{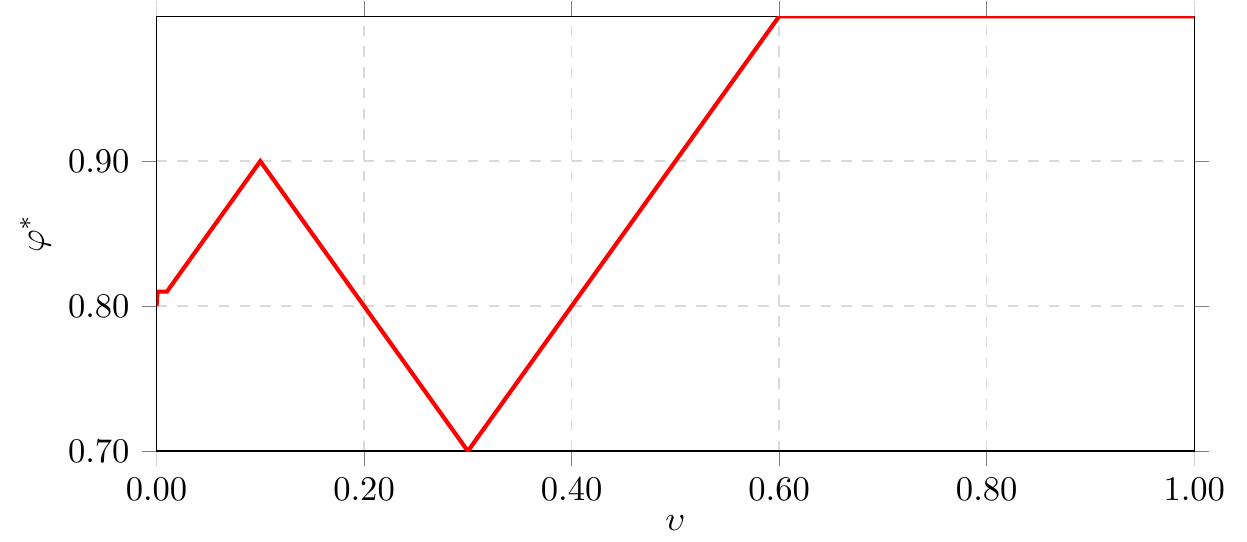}
  \caption{The behavior of $\varphi^*$ as a function of $\upsilon$ for $\pi_1 = 0.01$.}    
  \label{fig:phi_star_vs_upsilon}
\end{figure}

Figure~\ref{fig:phi_star_vs_upsilon} illustrates the exact behavior of $\varphi^*$ against $\upsilon$ for the case where $\phi_1 = 0$. As can be seen from the figure, $\varphi$ should be quite large. 

\begin{figure}
  \centering
  \includegraphics[width=5in]{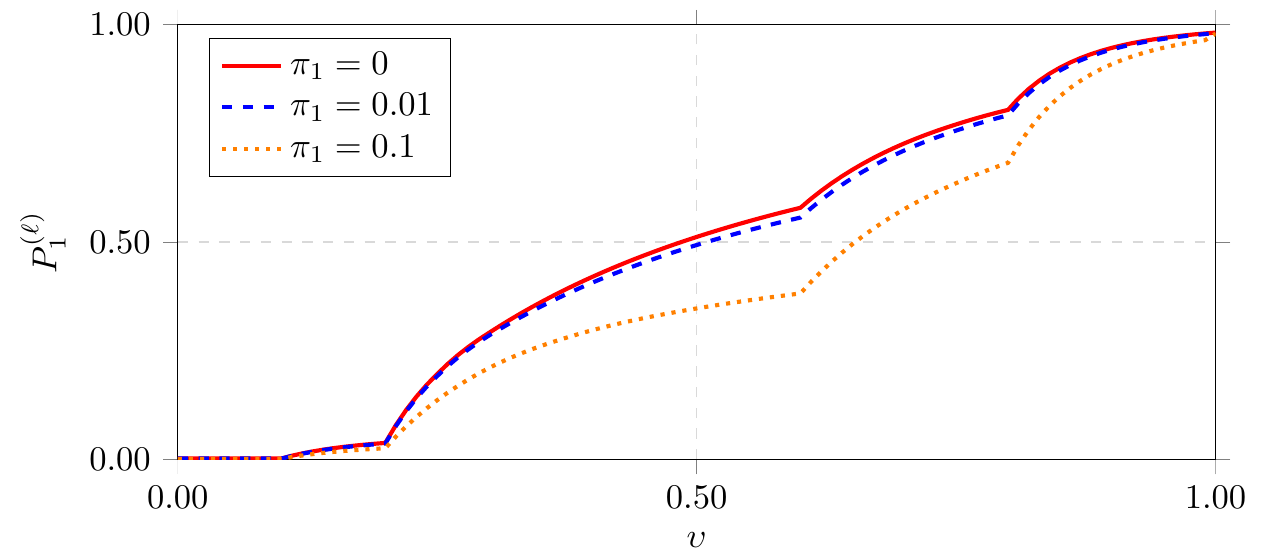}
  \caption{The optimum $P_{e_1}$ as a function of $\upsilon$ for different values $\pi_1$.}    
  \label{fig:pe_vs_upsilon}
\end{figure}

Figure \ref{fig:pe_vs_upsilon} illustrates $P_{e_1} = 1-P_{c_1}$ for the best chosen threshold, $\varphi^*$, as a function of $\upsilon$ for various choices of $\pi_1$.

\bibliographystyle{IEEEtran} 
\bibliography{abrv,conf_abrv,nn_lib}

\end{document}